\documentclass[twocolumn,letterpaper]{IEEEAerospaceCLS}  % only supports two-column, letterpaper format
\usepackage{times}
\usepackage{rawfonts}
\usepackage{geometry}
\usepackage{cite}
\usepackage{amssymb}
\usepackage{amsmath,bm}
\usepackage[]{graphicx,float,latexsym,amssymb,amsfonts,amsmath,amstext,times}
\usepackage{xcolor}
\definecolor{deeppink}{rgb}{1.0, 0.08, 0.58}
\usepackage{hyperref}
\hypersetup{
  colorlinks = true,  % colours link4.20s instead of ugly boxes
  urlcolor   = deeppink,  % colour for external hyperlinks
  linkcolor  = blue,  % colour of internal links
  citecolor  = red,  % colour of citations
  pdfview    = XYZ,  % page fit after navigation. Options: XYZ, Fit, FitH, FitV, FitB, FitR, FitBH, FitBV
}
\usepackage[]{graphicx}    % We use this package in this document
\newcommand{\ignore}[1]{}  % {} empty inside = %% comment

\begin{document}
\title{Compositional Diffusion Models for Powered Descent Trajectory Generation with Flexible Constraints}

\author{%
Julia Briden\\ 
Massachusetts Institute of Technology\\
77 Massachusetts Ave.\\
Cambridge, MA 02139\\
jbriden@mit.edu
\and 
Yilun Du\\
Massachusetts Institute of Technology\\
77 Massachusetts Ave.\\
Cambridge, MA 02139\\
yilundu@mit.edu\\
\and
Enrico M. Zucchelli\\
Massachusetts Institute of Technology\\
77 Massachusetts Ave.\\
Cambridge, MA 02139\\
enricoz@mit.edu
\and 
Richard Linares\\
Massachusetts Institute of Technology\\
77 Massachusetts Ave.\\
Cambridge, MA 02139\\
linaresr@mit.edu
%%%% IMPORTANT: Use the correct copyright information--IEEE, Crown, or U.S. government. %%%%%
\thanks{\footnotesize 979-8-3503-5597-0/25/$\$31.00$ \copyright2025 IEEE}              % This creates the copyright info that is the correct 2025 data.
%\thanks{{U.S. Government work not protected by U.S. copyright}}         % Use this copyright notice only if you are employed by the U.S. Government.
%\thanks{{979-8-3503-5597-0/25/$\$31.00$ \copyright2025 Crown}}          % Use this copyright notice only if you are employed by a crown government (e.g., Canada, UK, Australia).
%\thanks{{979-8-3503-5597-0/25/$\$31.00$ \copyright2025 European Union}}    % Use this copyright notice is you are employed by the European Union.
}

\maketitle

\thispagestyle{plain}
\pagestyle{plain}

\maketitle

\thispagestyle{plain}
\pagestyle{plain}

\begin{abstract}
This work introduces TrajDiffuser, a compositional diffusion-based flexible and concurrent trajectory generator for 6 degrees of freedom powered descent guidance. TrajDiffuser is a statistical model that learns the multi-modal distributions of a dataset of simulated optimal trajectories, each subject to only one or few constraints that may vary for different trajectories. During inference, the trajectory is generated simultaneously over time, providing stable long-horizon planning, and constraints can be composed together, increasing the model's generalizability and decreasing the training data required. The generated trajectory is then used to initialize an optimizer, increasing its robustness and speed.

\end{abstract}

\tableofcontents

%%%%%%%%%%%%%%%%%%%%%%%%%%%%%%%%%%%%%%
\section{Introduction}
%%%%%%%%%%%%%%%%%%%%%%%%%%%%%%%%%%%%%%
With the increasing need for autonomous systems for space applications, demand has grown for implementing advanced guidance, navigation, and control algorithms onboard space-grade processors. One critical technology is the use of direct optimization-based algorithms for trajectory optimization. While current research is working to increase the computational capabilities of radiation-hardened processors, long flight verification times make rapid hardware development difficult.
A guidance logic seeks to solve a two-point boundary value problem~(TBVP) to feasibly guide the spacecraft from its current state to the target state. The Apollo guidance algorithm analytically finds a polynomial trajectory that physically connects the initial and final conditions~\cite{cherry1964general,klumpp1974apollo}.
Because of five decades of advancements in hardware and software engineering, modern optimal guidance algorithms can now aim to seek increasingly complex trajectories that minimize propellant consumption while satisfying several constraints. These algorithms traditionally fall into two categories: direct and indirect methods. Direct methods solve for a discrete-time sequence of thrust vectors by formulating a large optimization problem. Indirect methods seek the initial costates satisfying Pontryagin's minimum principle~\cite{pontryagin2018mathematical} by root-finding. Convex formulations for the direct solution of the 3 degrees of freedom (DoF) powered descent guidance~\cite{acikmese2007convex} allow theoretical guarantees but lack flexibility. For the 6 DoF formulation, which is more exhaustive and precise, the state of the art is the successive convexification~(SCvx) algorithm~\cite{szmuk2018successivefinaltime,Szmuk2019}, which is more computationally demanding and cannot provide guarantees of convergence. Similarly, analytical solutions to the 3 DoF indirect problem can only be obtained after simplifications \cite{lu2012versatile,lu2018propellant,lu2023propellant}, but adapting these algorithms to satisfy more general constraints can be challenging.

Computational efficiency of direct and indirect methods alike can be increased by appropriately choosing an initial guess. A judicious choice of initial guess also decreases the non-convergence rate for non-convex problems. The process of generating an initial guess that is close to the solution is called ``warm-start". Recent research efforts have been directed towards using artificial intelligence~(AI)-generated initial guesses for applications where fast convergence is essential. This process involves a trade-off between ``offline" and ``online" computational demands, meaning that a large investment in computational cost before the mission allows immediate processing efficiency in-flight. These approaches generally involve the offline generation of a wide dataset of trajectories via optimization. Such dataset is then used to train AI models that can later quickly reproduce online a learned trajectory, adapted to the state observed by the spacecraft. The approximate trajectory then warm-starts the optimizer. The AI model can be used to generate entire sequences of states and control inputs, such as the deep neural network~(DNN) suggested by Cheng et al.~\cite{8587201} for solar sail optimal control, the transformer proposed by Guffanti et al.\cite{guffanti2024} for spacecraft rendezvous guidance, or the long short term memory~(LSTM) neural network and transformer neural network for lunar rover model predictive control and powered descent guidance developed by Briden et al. \cite{briden2023constraint}. Alternatively, the optimization process can be aided with less input from the AI. For example, Li and Gong~\cite{Li2022} use a DNN to predict just the final time of a powered descent trajectory, while Briden et al.~\cite{briden2024} use a transformer to guess which constraints are active during powered descent accurately, and Cauligi et al.~\cite{CauligiCulbertsonEtAl2022} use a DNN to map to a logical strategy which allows the practitioner to solve a mixed-integer convex program (MICP) as a convex optimization problem. These approaches accelerate the optimizer without generating an initial guess for the entire trajectory.

Many of the AI examples mentioned so far generate a trajectory sequentially. This can lead to cumulative errors and, in some cases, divergence. On the other hand, stability of the solution may be improved by generating an entire trajectory all at once. Diffusion models can be used to do exactly that. A diffusion model learns high-dimensional distributions from data and generates new samples by denoising~\cite{sohl2015deep,ho2020denoising}. Crucially, additional information can be provided to the models to condition the sampling. While diffusion models are mostly known for image generation from text~\cite{ramesh2022hierarchical,saharia2022photorealistic}, recent advancements have proposed diffusion models for trajectory generation and optimization in robotics~\cite{janner2022planning,ajay2022conditional}. In this framework a trajectory is represented as a matrix, thus a single-channel image, where column $i$ is the concatenated states and actions at time $t_i$. Note that while diffusion is an iterative process, the iterations are not over the time steps of the trajectory, which instead emerges concurrently. Further, diffusion models can learn multi-modal distributions, opposed to deterministic AI networks which only learn the mean. This can have catastrophic consequences when optimal trajectories overcome an obstacle equivalently from either side. Diffusion models have been proposed for ballistic trajectories generation~\cite{presser2024diffusion} and low-thrust cislunar transfer~\cite{li2024efficient}.

To allow flexiblity for online use, it would be ideal to have a model that can work with different sets of constraints, possibly without exponentially increasing the dataset size. Diffusion models can be composed together to draw samples from probability distributions never seen in training~\cite{du2020compositional,liu2022compositional}. For example, the set of trajectories satisfying multiple constraints can be sampled by composing the sets of all trajectories that satisfy only one of the desired constraints at a time. Note that composition requires independence between distributions, which is always true when combining sets with different constraints, because it is equivalent to seeking the intersection of sets. Composing constraints exponentially reduces the number of required training examples; only one constraint at a time needs to be satisfied during training, and then any combination of constraints can be obtained during sampling.
On the other hand, accuracy might be decreased compared to a model trained over all possible combinations of constraints; however, such performance reduction is for the most part due to the sampling method~\cite{du2023reduce}, and it can be circumvented using annealed Markov Chain Monte Carlo~(MCMC)~\cite{neal2001annealed}.

In this work, we create TrajDiffuser, a trajectory design tool that provides efficient and generalizable initial guesses for computationally expensive trajectory optimization tasks through compositional diffusion, enabling dynamically feasible trajectory generation on current flight-grade processors. The main features of TrajDiffuser are 1) concurrent trajectory planning opposed to sequential planning, 2) multi-modality as opposed to averaging, and 3) compositionality. In turn, these features provide the following advantages to the trajectory generation tool: 1) increased stability of the generated trajectory, 2) ability to avoid obstacles from both sides, and 3) increased flexibility with reduced training time.
The major contributions of this paper are:
\begin{itemize}
    \item The development of TrajDiffuser, a six degree-of-freedom diffusion-based flexible trajectory generator that provides warm starts for powered descent guidance problems with several combinations of constraints not seen during training.
    \item The derivation of a bound on the sample accuracy of TrajDiffuser.
    \item Empirical testing of the performance of TrajDiffuser in simulation, illustrating its efficacy as warm-start for powered descent guidance problems.
    \item The development of C-TrajDiffuser, Composable TrajDiffuser, which enables generalizable trajectory generation during inference time via product, mixture, and negation operations. Test cases include state-triggered constraints and composition with a lower-fidelity drag model.
\end{itemize}

%%%%%%%%%%%%%%%%%%%%%%%%%%%%%%%%%%%%%%%%%%%
\subsection{Paper Organization}
%%%%%%%%%%%%%%%%%%%%%%%%%%%%%%%%%%%%%%%%%%%

This work is organized as follows. Section \ref{sec:background} reviews the diffusion modeling process and its formulation as a specific type of energy-based model. Utilizing this extension, Section \ref{sec:compdiff} defines product, mixture, and negation formulations for the composition of energy functions. Additionally, an upper bound on the magnitude of the error between the score functions for true target product distribution and the reverse diffusion process approximated product distribution is derived in Section \ref{sec:compdiff}. Section \ref{sec:sixdof} defines the six-degree-of-freedom minimum time powered descent guidance problem and the successive convexification (SCvx) process used to generate samples for the training of TrajDiffuser. Section \ref{sec:constraints} describes the state-triggered constraint formulation, including the velocity-triggered angle of attack constraint used for composition in this work. Section \ref{sec:experiments} includes the training process, architecture, and performance benchmarking for TrajDiffuser. Runtime benchmarking for TrajDiffuser, numerical optimizing-based sampling, the formulation and evaluation of the product and negation state-triggered constraints, and product composition for introducing drag into TrajDiffuser are also included in Section \ref{sec:experiments}. 
%Section \ref{sec:discussion} discusses the implications of the experiments and provides directions for future work. 
Conclusions from this work are included in Section \ref{sec:conclusions}.

%%%%%%%%%%%%%%%%%%%%%%%%%%%%%%%%%%%%%%%%%%%%%
\section{Background}
\label{sec:background}
%%%%%%%%%%%%%%%%%%%%%%%%%%%%%%%%%%%%%%%%%%%%%

\subsection{Diffusion Models}

\begin{figure}
    \centering
    \includegraphics[width=\linewidth]{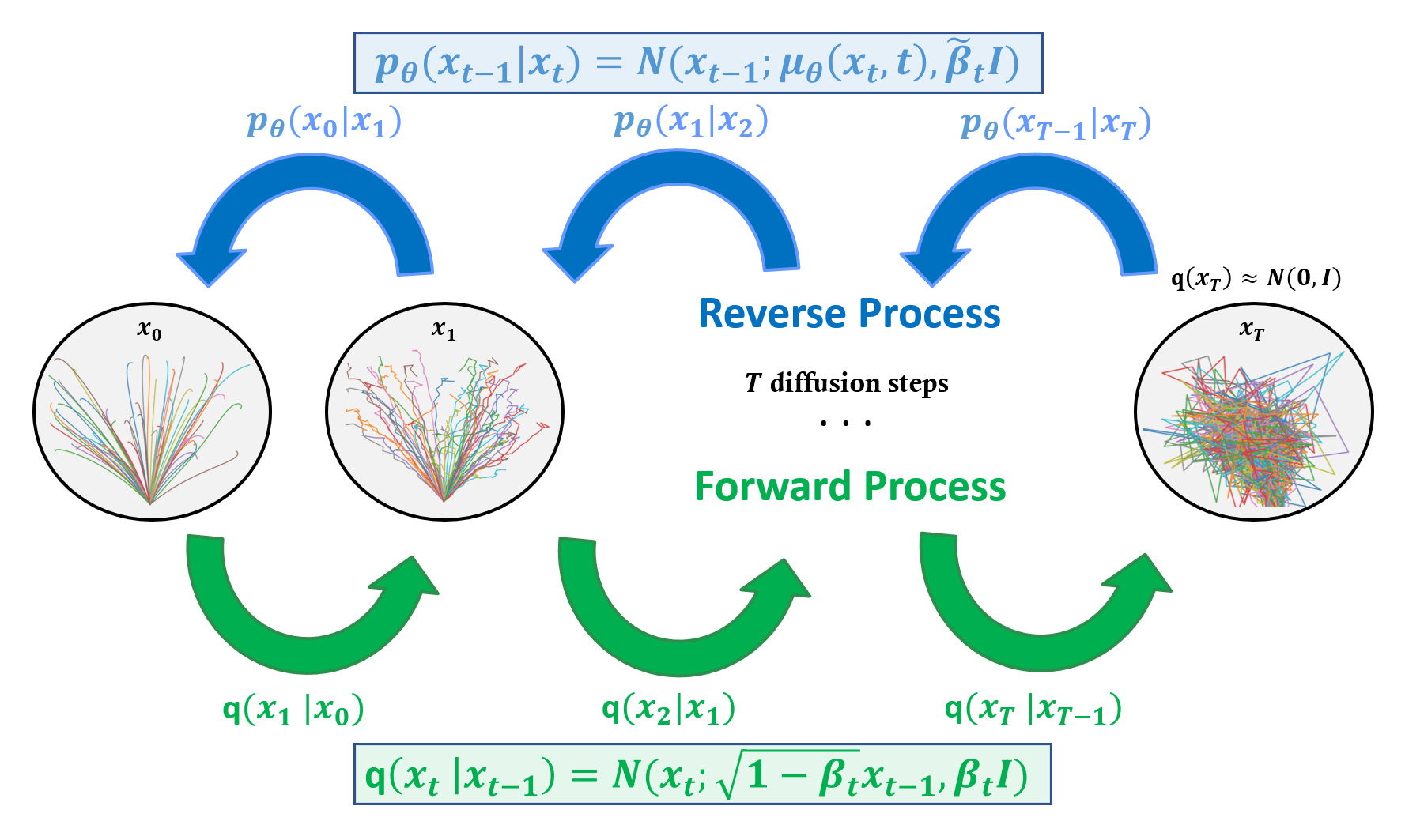}
    \caption{The forward and reverse processes in Gaussian diffusion.}
    \label{fig:forward_reverse_process}
\end{figure}

To learn the distribution $q({{\bm{x}}}_0)$ that generates trajectories ${\bm{\tau}} \sim q({\bm{x}}_0)$, Gaussian diffusion treats the direct learning of the data's probability distribution as a denoising problem. Diffusion models are Markov chains consisting of two processes: 1) the forward, or noising, process $q({\bm{x}}_t |{\bm{x}}_{t-1})$ and 2) the reverse, or denoising, process $p_{\bm{\theta}} ({\bm{x}}_{t-1} | {\bm{x}}_t)$, where ${\bm{x}}_1, \dots, {\bm{x}}_T$ are latent variables generated from the Markov process ${\bm{x}}_t \sim q({\bm{x}}_t | {\bm{x}}_{t-1})$, and ${\bm{\theta}}$ is a set of parameters to be learned. The two processes are depicted in Figure~\ref{fig:forward_reverse_process}.

\begin{enumerate}
    \item The \textbf{forward process} $q({\bm{x}}_t |{\bm{x}}_{t-1})$ transitions the original trajectory data by first scaling ${\bm{x}}_{t-1}$ by $\sqrt{1-\beta_t}$ and then adding Gaussian noise with variance $0 < \beta_t \leq 1$. This use of transitions between auxiliary noisy trajectories allows $q({\bm{x}}_T)$ to approach the standard normal distribution $q({\bm{x}}_t) \rightarrow \mathcal{N}$$( 0, I )$ as $t \rightarrow T$, for a sufficiently large $T$.

    \item The \textbf{reverse process} $p_{\bm{\theta}} ({\bm{x}}_{t-1} | {\bm{x}}_t)$ is also defined by a Gaussian distribution for sufficiently small variance $\beta_t$ \cite{sohl2015}. Given the learned perturbation, ${\bm{\epsilon}}_{\bm{\theta}} ({\bm{x}}_t, t)$, the conditional probability for the reverse process is $p_{\bm{\theta}} ({\bm{x}}_{t-1} | {\bm{x}}_t) = {{\mathcal{N}}} ({\bm{x}}_{t-1} ; {\bm{\mu}}_{\bm{\theta}} ({\bm{x}}_t, t), \tilde \beta_t I)$, with the mean defined as:

    \begin{equation}
        {{{\bm{\mu}}}}_{\bm{\theta}} ({\bm{x}}_t, t) = \frac{1}{\sqrt{\alpha_t}} \left({\bm{x}}_t - \frac{\beta_t}{\sqrt{1 - \bar \alpha_t}} {\bm{\epsilon}}_{\bm{\theta}} ({\bm{x}}_t, t)\right),
    \label{eq: mean}
    \end{equation}

    where $\alpha_t = \Pi_{t=1}^T \beta_t$, the variance of the reversal is $\tilde \beta_t = \frac{1 - \bar \alpha_{t-1}}{1 - \bar \alpha_t}$, and $\bar \alpha_t = \Pi_{t=1}^T(1-\beta_t)$.
\end{enumerate}

\subsubsection{Training}

To train the neural network parametric model to predict ${\bm{\epsilon}}_{\bm{\theta}}$, the analytic expression for sampling at each diffusion step $t$ can be used:
\begin{equation}
    {\bm{x}}_t ({\bm{x}}_0, {\bm{\epsilon}}) = \sqrt{1-\sigma_t^2} {\bm{x}}_0 + \sigma_t {\bm{\epsilon}},
\end{equation}
where $\bm{\epsilon} \sim {\mathcal{N}} (0, I)$ and $\sigma_t = \sqrt{1 - \bar \alpha_t}$. Generally, the goal of diffusion modeling is to maximize the following variational bound on the marginal likelihood for the reverse process~\cite{ho2020denoising}:
\begin{align}
    {{\mathbb{E}}}\left[\log p_{\bm{\theta}} ({\bm{x}}_0)\right] &\geq {\mathbb{E}}_{q({\bm{x}}_1, \dots, {\bm{x}}_T | {\bm{x}}_0)} 
    \left[ \log p_{\bm{\theta}} ({\bm{x}}_0 | {\bm{x}}_1) \right. \nonumber \\
    &\quad + \sum_{t=1}^T D_{KL} \left( q({\bm{x}}_t | {\bm{x}}_{t-1}, {\bm{x}}_0) \| p({\bm{x}}_t | {\bm{x}}_{t-1}) \right) \nonumber \\
    &\quad \left. + D_{KL} \left( q({\bm{x}}_T | {\bm{x}}_0) \| p({\bm{x}}_T) \right) \right].
\end{align}
Given that $\log p_{\bm{\theta}} ({\bm{x}}_0 | {\bm{x}}_1)$ is not in terms of $q$ and $ D_{KL} (q({\bm{x}}_T | {\bm{x}}_0) || p({\bm{x}}_T)) \approx 0$ since the forward process $q({\bm{x}}_T | {\bm{x}}_0)$ converges to $p({\bm{x}}_T)$, both the first and last terms cancel out. This leaves the following term remaining:
\begin{align}
    {\mathbb{E}}\left[\log p_{\bm{\theta}} ({\bm{x}}_0)\right] \geq& {\mathbb{E}} \left[ \sum_{t=1}^T D_{KL} \left( q({\bm{x}}_t | {\bm{x}}_{t-1}, {\bm{x}}_0) \| p({\bm{x}}_t | {\bm{x}}_{t-1}) \right) \right] \\
    &= \sum_{t=1}^T C_t {\mathbb{E}}_{q({\bm{x}}_0), {\mathcal{N}}({\bm{\epsilon}} ; 0, I)} \left[ \| {\bm{\epsilon}} - {\bm{\epsilon}}_{\bm{\theta}} ({\bm{x}}_t, t) \|^2 \right],
\end{align}
with the time-dependent constant $C_t$. Therefore, for training the neural network to predict ${\bm{\epsilon}}_{\bm{\theta}}$ at diffusion step $t$, the lower bound on marginal likelihood is often used as the loss function, with $C_t = 1$ unless otherwise specified:
\begin{equation}
    {\mathcal{L}}_t ({\bm{\theta}}) = {\mathbb{E}}_{q({\bm{x}}_0) {\mathcal{N}} ({\bm{\epsilon}}; 0, I) } \left[ \| {\bm{\epsilon}} - {\bm{\epsilon}}_{\bm{\theta}} ({\bm{x}}_t ({\bm{x}}_0, {\bm{\epsilon}}), t) \|^2 \right].
    \label{eqn:loss_diffusion}
\end{equation}
The loss for the full diffusion process can then be obtained by summing over the $T$ steps ${\mathcal{L}} ({\bm{\theta}}) = \sum_{t=1}^T {\mathcal{L}}_t ({\bm{\theta}})$.

\subsubsection{Sampling}

Once trained, ${\bm{\epsilon}}_{\bm{\theta}}$ can be substituted into Equation \eqref{eq: mean} to compute the mean of each reverse step, obtaining the conditional probability:

\begin{equation}
    p_{\bm{\theta}} ({\bm{x}}_t | {\bm{x}}_{t+1}) = {\mathcal{N}} \left( {\bm{x}}_t ; {\bm{\mu}}_{\bm{\theta}} ({\bm{x}}_{t-1}, t), {\tilde \beta}_t I \right),
\end{equation}

where the variance of the reversal is $\tilde \beta_t = \frac{1 - \bar \alpha_{t-1}}{1 - \bar \alpha_t}$ and $\bar \alpha_t = \Pi_{t=0}^T(1-\beta_t)$.

Sampling using reverse diffusion consists of two steps: 1) drawing one or more samples ${\bm{x}}_T \sim p({\bm{x}}_T) \approx \mathcal{N}$$( 0, I )$ and 2) sampling
\begin{equation}
    {\bm{x}}_{t-1} = {\bm{\mu}}_{\bm{\theta}} ({\bm{x}}_{t}, t) + \sqrt{\tilde \beta_t} {\bm{\epsilon}},
    \label{eqn:diffusion_sample}
\end{equation}
where ${\bm{\epsilon}} \sim \mathcal{N}$$( 0, I )$, for $T$ diffusion steps.

\subsection{Diffusion Models as Energy-Based Models}
\label{sect:diffusion_ebm}

An alternative view of diffusion models is to view the denoising function ${\bm{\epsilon}}_{\bm{\theta}}(\bm{x}_t, t)$ conditioned at each of the T timesteps, also referred to as the {\it score function}, as implicitly parameterizing a sequence of $T$ energy functions ${\bm{\epsilon}}_{\bm{\theta}}(\bm{x}_t, t) = \frac{\nabla E_{\bm{\theta}}(\bm{x}_t, t)}{\sqrt{1 - \bar \alpha_t}}$~\cite{du2023reduce}.  Each energy function corresponds to a EBM~\cite{lecun2006tutorial,du2019implicit}, where $p_{\bm{\theta}}^t(\bm{x}) \propto e^{-E_{\bm{\theta}}(\bm{x}, t)}$. In this interpretation, $p_{\bm{\theta}}^T(\bm{x})$, corresponds to the denoising function conditioned at the highest noise level timestep, and is the Gaussian distribution $\mathcal{N}$$(0, 1)$ while $p_{\bm{\theta}}^1(\bm(x))$ is the ground truth data distribution 
 of samples $p_D\bm(x)$~\cite{du2023reduce}. 

The sampling procedure for diffusion models in Equation~\ref{eqn:diffusion_sample} can be written in terms of ${\bm{\epsilon}}_{\bm{\theta}}(\bm{x}_t, t)$ as
\begin{equation}
    {\bm{x}}_{t-1} = \frac{1}{\sqrt{\alpha_t}} \left({\bm{x}}_t - \frac{\beta_t}{\sqrt{1 - \bar \alpha_t}} {\bm{\epsilon}}_{\bm{\theta}} ({\bm{x}}_t, t)\right) + \sqrt{\tilde \beta_t} {\bm{\epsilon}},
\end{equation}
where by substituting ${\bm{\epsilon}}_{\bm{\theta}}(\bm{x}_t, t) = \frac{\nabla E_{\bm{\theta}}(\bm{x}_t, t)}{\sqrt{1 - \bar \alpha_t}}$, we obtain an expression in terms of the energy function $E_{\bm{\theta}}(\bm{x}_t, t)$,
\begin{equation}
    {\bm{x}}_{t-1} = \frac{1}{\sqrt{\alpha_t}} \left ({\bm{x}}_t - \beta_t  \nabla E_{\bm{\theta}}(\bm{x}_t, t) + \sqrt{\tilde \beta_t'} {\bm{\epsilon}} \right ).
    \label{eqn:ebm_sample}
\end{equation}
The expression in Equation~\ref{eqn:ebm_sample} inside the $\frac{1}{\sqrt{\alpha_t}}$ contraction expression corresponds to unadjusted Langevin dynamics~(ULA) MCMC sampling step~\cite{roberts1996exponential,du2019implicit} in the EBM distribution $p_{\bm{\theta}}^t(\bm{x})$, while the contraction expression  $\frac{1}{\sqrt{\alpha_t}}$ is scaling factor that transitions a sample from distribution $p_{\bm{\theta}}^t(\bm{x})$ to $p_{\bm{\theta}}^{t-1}(\bm{x})$. We can therefore view the diffusion process as sampling sequentially using MCMC on a sequence of EBM distributions starting from Gaussian noise ($p_{\bm{\theta}}^T(\bm{x})$) to $p_{\bm{\theta}}^1(\bm(x))$. This EBM perspective gives us a method to compose the distributions specified by multiple diffusion models together by directly combining the intermediate EBM distributions $p_{\bm{\theta}}^t(\bm{x})$ specified by each model together. 

To increase the number of ways intermediate distributions $p_{\bm{\theta}}^t(\bm{x})$ can be combined, we train diffusion models to model an EBM $E_{\bm{\theta}}(\bm{x}_t, t)$ as opposed to the denoising/score function ${\bm{\epsilon}}_{\bm{\theta}}(\bm{x}_t, t)$.  To train each EBM, we directly supervise $\nabla_{\bm{x}}E_{\bm{\theta}}(\bm{x}_t, t)$ with the denoising objective of Equation~\ref{eqn:loss_diffusion} and differentiate through this gradient similar to prior work on compositional models~\cite{du2023reduce}. This EBM parameterization of diffusion models allows us to compose distributions following a set of compositional operations~\cite{du2020compositional} which we discuss in Section~\ref{sec:compdiff}.

% the direction of the gradient of the energy, which is also called the score, where . This interpretation enables us to take two distributions, represented as 

\section{Compositional Diffusion}
\label{sec:compdiff}

A current issue with leveraging learned models for efficient initial guess generation for trajectory optimization is the lack of generalizability to new or even slightly different scenarios; changing only one constraint in the optimization problem would require training an entirely new learned model. Due to the structure of diffusion models as distributions, compositional generation can be utilized to obtain samples from entirely new models by composing learned ``building block" distributions together without the need for training.

In particular, we use the connection of diffusion models and EBMs discussed in Section~\ref{sect:diffusion_ebm}. This enables us to form new distributions by composing multiple energy functions together. The composition of energy functions enables the following operations~\cite{du2020compositional}:

\begin{enumerate}
    \item \textbf{Products:} The intersection of $N$ distributions $\cap_{i=1}^N q^i({\bm{x}})$ is roughly equivalent to the product of all $N$ distributions,
    \begin{equation}
        q_\text{prod} ({\bm{x}}) = \frac{1}{Z} \Pi_{i=1}^N p^i_{\bm{\theta}} ({\bm{x}}, t) \propto e^{\sum_{i=1}^N -E^i_{\bm{\theta}} ({\bm{x}},t)},
        \label{eq: prod}
    \end{equation}
    where $Z = \int \Pi_{i=1}^N p^i_{\bm{\theta}} ({\bm{x}}, t) dx$ is the normalization constant.

    \item \textbf{Mixtures:} The union of $N$ distributions $\cup_{i=1}^N q^i({\bm{x}})$ can be formulated as a mixture or sum of all $N$ distributions,
    \begin{equation}
        q^{\text{mix}} ({\bm{x}}) = \frac{1}{N} \sum_{i=1}^N p^i_{\bm{\theta}} ({\bm{x}}, t) \propto \frac{1}{N} \sum_{i=1}^N e^{-E^i_{\bm{\theta}} ({\bm{x}},t)}.
        \label{eq: mix}
    \end{equation}
    where $\alpha$ controls the degree $p_1({\bm{x}})$ is inverted.

    \item \textbf{Negation: } The negation of a distribution $q^0 ({\bm{x}})$ by another distribution $q^1 ({\bm{x}})$ can be achieved by inverting the density $q^1 ({\bm{x}})$ with respect to $q^0 ({\bm{x}})$,
    \begin{equation}
        q^{\text{neg}} ({\bm{x}}) \propto \frac{p^0_{\bm{\theta}} ({\bm{x}}, t)}{p^1_{\bm{\theta}} ({\bm{x}}, t)^\alpha} \propto e^{E^1_{\bm{\theta}} ({\bm{x}},t)^\alpha - E^0_{\bm{\theta}} ({\bm{x}},t)},
        \label{eq: neg}
    \end{equation}
\end{enumerate}

The product, mixture, and negation-based compositional structures allow completely generalized and complex models to be constructed from relatively simple and low-cost trained diffusion models or PDFs. For example, a dynamically-feasible diffusion model could be composed with pre-trained models for specific state and control constraints based on mission requirements. In the following section, we will cover the options for sampling from the composed distributions, including time complexity and accuracy trade-offs.

\subsection{Reverse Diffusion Sampling for Compositional Diffusion}

The primary challenge with directly sampling from the composed PDFs defined in Equations \eqref{eq: prod}-\eqref{eq: neg} is the assumption that score functions can be directly combined to obtain the target distribution; Du et al. prove that sampling using the score function from the target product distribution:
\begin{align}
    \tilde {\bm{\epsilon}}_{{\bm{\theta}} \; \text{prod}} ({\bm{x}},t) &= {\bm{\nabla}}_{\bm{x}} \log \tilde q_\text{prod} ({\bm{x}},t) \\
    &= {\bm{\nabla}}_{\bm{x}} \log \Pi_{i=1}^N p^i_{\bm{\theta}} ({\bm{x}}, t) \\
    &= {\bm{\nabla}}_{\bm{x}} \log \left(\int \Pi_{i=1}^N p^i_{\bm{\theta}} ({\bm{x}}_0) q({\bm{x}}_t | {\bm{x}}_0) d {\bm{x}}_0\right)
\end{align}
is not equivalent to the reverse diffusion process of sampling from the composed scores of multiple models for $t > 0$:
\begin{align}
    {\bm{\epsilon}}_{{\bm{\theta}} \; \text{prod}} ({\bm{x}},t) &= {\bm{\nabla}}_{\bm{x}} \log q_\text{prod} ({\bm{x}},t) \\
    &= \sum_{i=1}^N {\bm{\nabla}}_{\bm{x}} \log \left(\int p^i_{\bm{\theta}} ({\bm{x}}_0) q({\bm{x}}_t | {\bm{x}}_0) d {\bm{x}}_0\right) \\
    &= \sum_{i=1}^N {\bm{\epsilon}}_{\bm{\theta}}^i.
\end{align}
Reverse diffusion on ${\bm{\epsilon}}_{{\bm{\theta}} \; \text{prod}} ({\bm{x}},t)$ is performed in practice, instead of using the true target distribution, $\tilde {\bm{\epsilon}}_{{\bm{\theta}} \; \text{prod}} ({\bm{x}},t)$, since the score is easy to compute using a sequence of distributions which smoothly interpolate between $q_\text{prod}$ and ${\mathcal{N}}(0, I/2)$ \cite{du2023reduce}. MCMC sampling methods can be used instead, but they come with a time complexity vs. accuracy trade-off, surpassing the time complexity of the efficient reverse process sampling in diffusion. Since the primary purpose of this framework is to provide computationally efficient initial guesses to the numerical optimizer, reverse diffusion is used for sampling in this work to achieve trajectory samples that are biased toward composed constraint satisfaction. To further understand the sampling accuracy degradation from using the reverse diffusion formulation, we can bound the difference between the true product distribution and the reverse diffusion sampling process of the sum of score functions as follows,
\begin{align}
    \delta_{\text{prod}} ({\bm{x}},t) &= |{\bm{\epsilon}}_{{\bm{\theta}} \; \text{prod}} ({\bm{x}},t) - \tilde {\bm{\epsilon}}_{{\bm{\theta}} \; \text{prod}} ({\bm{x}},t)|.
\end{align}
The bound on this difference, derived in Appendix \ref{sec: appendix product}, can be defined as the following:
\begin{equation}
    \delta_{\text{prod}} ({\bm{x}},t) 
    \leq \frac{1}{1-\bar \alpha_t} ((N-1)\|{\bm{x}}_t\| + \sqrt{\bar \alpha_t} \| \Delta {\bm{\mu}} \|).
\end{equation}
Where $\bar \alpha = \Pi_{t=0}^T (1- \beta_t)$, $N$ is the number of composed models, $\|{\bm{x}}_t\|$ is the norm of the diffusion sample at diffusion step $t$, and $\Delta {\bm{\mu}} = N ( {\mathbb{E}}_{\tilde q_{\text{prod}} ({\bm{x}}_0 | {\bm{x}}_t)} [{\bm{x}}_0] - \bar {\bm{\mu}})$ with $\bar {\bm{\mu}} = \frac{1}{N} \sum_{i=1}^N {\mathbb{E}}_{q^i ({\bm{x}}_0 | {\bm{x}}_t)} [{\bm{x}}_0]$. We can sum over $T$ diffusion steps to get the final bound for the full reverse process:
\begin{equation}
    \delta_{\text{prod}} ({\bm{x}}) 
    \leq \sum_{t=1}^T \frac{1}{1-\bar \alpha_t} ((N-1)\|{\bm{x}}_t\| + \sqrt{\bar \alpha_t} \| \Delta {\bm{\mu}} \|).
\end{equation}
Finally using our knowledge of the diffusion process and assuming ${{\text{Var}}_{q^i}} ({\bm{x}}_0) = {{\text{Var}}_{q^j}} ({\bm{x}}_0)$ for $i, j \in 1, \dots N$, we have the bounds $\| {\bm{x}}_t \| \leq \sqrt{\bar \alpha_t {\text{Tr}} ({\text{Var}} ({\bm{x}}_0)) + (1-\bar{\alpha}_t) d}$ and $\| \Delta {\bm{\mu}} \| \leq {\sqrt{ 2 {\text{Tr}} ( {{\text{Var}}_{q^0}} ({\bm{x}}_0)})}$, giving us the following bound in terms of variance:
\begin{align}
    \delta_{\text{prod}} ({\bm{x}}) 
    &\leq \sum_{t=1}^T \frac{1}{1-\bar \alpha_t} ((N-1) \sqrt{\bar \alpha_t {\text{Tr}} ({\text{Var}} ({\bm{x}}_0)) + (1-\bar{\alpha}_t) d}  \\
    &+ N \sqrt{\bar \alpha_t} {\sqrt{ 2 {\text{Tr}} ( {{\text{Var}}_{q^0}} ({\bm{x}}_0)})} ),
    \label{eq: bound}
\end{align}
where $d$ is the dimensionality of ${\bm{x}}_t$. The reverse diffusion error bound scales proportionally with an increasing number of models composed, $N$. Note that $\bar \alpha = \Pi_{t=0}^T (1- \beta_t) \in [0, 1)$, so $\bar \alpha$ will not make the bound infinite but a larger forward process variance $\beta_t$ will reduce the upper bound.
Reducing the number of diffusion steps $T$ may also keep $\bar \alpha$ from increasing the error bound if $\bar \alpha \in (0, 1)$.
Finally, data variances ${\text{Var}} ({\bm{x}}_0)$ and ${{\text{Var}}}_{q^0} ({\bm{x}}_0)$ directly control the upper bound as well, making normalization procedures potentially useful if significant compositional errors are obtained.
This compositional bound will be plotted in our results to show empirically how tight reverse diffusion results hold.

\section{Six Degrees-of-Freedom Powered Descent Guidance}
\label{sec:sixdof}

The 6 DoF free-final-time powered descent guidance problem formulation used in this work assumes that speeds are sufficiently low such that planetary rotation and changes in the planet's gravitational field are negligible. The spacecraft is assumed to be a rigid body with a constant center of mass and inertia and a fixed center of pressure. The propulsion consists of a single rocket engine that can be gimbaled symmetrically about two axes bounded by a maximum gimbal angle $\delta_\text{max}$. The engine is assumed to be throttleable between $T_\text{min}$ and $T_\text{max}$, remaining on until the terminal boundary conditions are met. The minimum-time 6 DoF powered descent guidance problem is formulated as follows:

\textbf{Cost Function:}
\begin{equation}
    \min_{t_f, \bm{T}_{{\mathcal{B}}(t)}} t_f
    \label{eq:cost}
\end{equation}
\textbf{Boundary Conditions:}
\begin{subequations}
\label{eq:boundary_conditions1}
\begin{align}
    & m(0) = m_{\text{wet}}\\
    & {\bm{r}}_{\mathcal{I}}(0) = r_{{\mathcal{I}}, i}  \\
    & {\bm{v}}_{\mathcal{I}}(0) = v_{{\mathcal{I}}, i}  \\
    & {\bm{\omega}}_{\mathcal{B}}(t_{0}) = {\bm{\omega}}_{{\mathcal{B}},i}\\
    & {\bm{r}}_{\mathcal{I}}(t_f) = {\bm{0}}\\
    & {\bm{v}}_{\mathcal{I}}(t_f) = {\bm{v}}_{{\mathcal{I}},f} \\
    & {\bm{q}}_{{\mathcal{B}} \leftarrow {\mathcal{I}}}(t_f) = {\bm{q}}_{{\mathcal{B}} \leftarrow {\mathcal{I}},  f}\\
    & {\bm{\omega}}_B(t_f) = 0 \\
    & {\bm{e_2}} \cdot {\bm{T}}_{\mathcal{B}} (t_f) = {\bm{e_3}} \cdot {\bm{T}}_{\mathcal{B}} (t_f) = 0
\end{align}
\end{subequations}
\textbf{Dynamics:}
\begin{subequations}
\label{eq:dynamics2}
\begin{align}
    & \dot{m}(t) = -\alpha_{\dot{m}} \|{\bm{T}}_{\mathcal{B}}(t) \|_2 \\
    & \dot{\bm{r}}_{\mathcal{I}}(t) = {\bm{v}}_{\mathcal{I}}(t)\\
    & \dot{\bm{v}}_{\mathcal{I}}(t) = \frac{1}{m(t)} {\bm{C}}_{{\mathcal{I}} \leftarrow {\mathcal{B}}}(t) {\bm{T}}_{\mathcal{B}}(t) + {\bm{g}}_{\mathcal{I}} \\
    & \dot{\bm{q}}_{{\mathcal{B}} \leftarrow {\mathcal{I}}}(t) = \frac{1}{2} \Omega_{{\bm{\omega}}_{\mathcal{B}}(t)} \bm{q}_{{\mathcal{B}} \leftarrow {\mathcal{I}}}(t) \\
    & {\bm{J}}_{\mathcal{B}} \dot{{\bm{\omega}}}_{\mathcal{B}}(t) = {\bm{r}}_{{\text{T}},{\mathcal{B}}} \times {\bm{T}}_{\mathcal{B}}(t) - {\bm{\omega}}_{\mathcal{B}}(t) \times {\bm{J}}_{\mathcal{B}} {\bm{\omega}}_{\mathcal{B}}(t)
\end{align}
\end{subequations}
\textbf{State Constraints:}
\begin{subequations}
\label{eq:state_constraints1}
\begin{align}
    & m_{\text{dry}} \leq m(t)
\end{align}
\end{subequations}
\textbf{Control Constraints:}
\begin{subequations}
\label{eq:control_constraints1}
\begin{align}
    & 0 < T_{\text{min}} \leq \|{\bm{T}}_{\mathcal{B}}(t)\|_2 \leq T_{\text{max}} \\
    & \cos \delta_{\text{max}} \|{\bm{T}}_{\mathcal{B}}(t)\|_2 \leq {\bm{e}}_1 \cdot {\bm{T}}_{\mathcal{B}}(t),
\end{align}
\end{subequations}

where the objective, Equation \eqref{eq:cost}, is to minimize the final time \( t_f \) while controlling the thrust vector \( \bm{T}_{\mathcal{B}}(t) \), which is expressed in the body frame \( \mathcal{B} \). The boundary conditions, Equation \eqref{eq:boundary_conditions1}, specify the initial and final states, where \( m(0) = m_{\text{wet}} \) represents the initial wet mass of the vehicle, \( {\bm{r}}_{\mathcal{I}}(0) = r_{{\mathcal{I}}, i} \) and \( {\bm{v}}_{\mathcal{I}}(0) = v_{{\mathcal{I}}, i} \) are the initial position and velocity in the inertial frame \( \mathcal{I} \), and \( \bm{\omega}_{\mathcal{B}}(t_0) = \bm{\omega}_{{\mathcal{B}}, i} \) denotes the initial angular velocity in the body frame. At the final time, the vehicle reaches the desired final position \( \bm{r}_{\mathcal{I}}(t_f) = \bm{0} \), final velocity \( \bm{v}_{\mathcal{I}}(t_f) = \bm{v}_{{\mathcal{I}}, f} \), and orientation described by the quaternion \( \bm{q}_{\mathcal{B} \leftarrow \mathcal{I}}(t_f) = \bm{q}_{{\mathcal{B}} \leftarrow {\mathcal{I}}, f} \), with the angular velocity \( \bm{\omega}_{\mathcal{B}}(t_f) = 0 \). The thrust direction constraints ensure that the thrust components along certain axes are zero at the final time. The dynamics, Equation \eqref{eq:dynamics2}, are governed by the mass depletion rate \( \dot{m}(t) \), position evolution \( \dot{\bm{r}}_{\mathcal{I}}(t) \), velocity dynamics \( \dot{\bm{v}}_{\mathcal{I}}(t) \), quaternion kinematics \( \dot{\bm{q}}_{{\mathcal{B}} \leftarrow {\mathcal{I}}}(t) \), and angular velocity evolution governed by the moment of inertia \( \bm{J}_{\mathcal{B}} \) and the applied thrust. Equations \eqref{eq:state_constraints1} and \eqref{eq:control_constraints1} ensure that the mass is bounded, a maximum tilt angle is not exceeded, and thrust is bounded.

To solve the non-convex minimization problem posed in Equations \eqref{eq:cost}-\eqref{eq:control_constraints1}, a series of convex second-order cone programming (SOCP) problems can be solved until convergence in a process known as successive convexification (SCvx). If the non-convex problem admits a feasible solution, the converged solution will be constraint-satisfying. The SOCP sub-problem that is iteratively solved during successive convexification is defined according to

\textbf{Cost Function:}
\begin{equation}
    \min_{\sigma^i, {\bm{u}}_k^i} \sigma^i + {\bm{\omega}}_\nu \|{\bm{\bar \nu}}^i\|_1 + {\bm{\omega}}_{{\bm{\Delta}}}^i \| {\bm{\bar {\bm{\Delta}}}}^i \|_2 + {\bm{\omega}}_{{\bm{\Delta}}_{\sigma}} \| {\bm{\Delta}_{\sigma}} \|_1
    \label{eq:cost_function}
\end{equation}
\textbf{Boundary Conditions:}
\begin{subequations}
\label{eq:boundary_conditions2}
\begin{align}
    & m_0^i = m_{\text{wet}} \\
    & {\bm{r}}_{{\mathcal{I}}, 0}^i = {{\bm{r}}}_{{\mathcal{I}}, i} \\
    & {\bm{v}}_{{\mathcal{I}}, 0}^i = {{\bm{v}}}_{{\mathcal{I}}, i} \\
    & {\bm{\omega}}_{{\mathcal{B}}, 0}^i = {\bm{\omega}}_{{\mathcal{B}}, i} \\
    & {\bm{r}}_{{\mathcal{I}}, K}^i = 0 \\
    & {\bm{v}}_{{\mathcal{I}}, K}^i = {\bm{v}}_{{\mathcal{I}}, f} \\
    & {\bm{q}}_{{\mathcal{B}} \leftarrow {\mathcal{I}}, K}^i = {\bm{q}}_{{\mathcal{B}} \leftarrow {\mathcal{I}}, f} \\
    & {\bm{\omega}}_{{\mathcal{B}}, K}^i = 0 \\
    & {\bm{e}}_2 \cdot {\bm{u}}_{{\mathcal{B}}, K}^i = {\bm{e}}_3 \cdot {\bm{u}}_{{\mathcal{B}}, K}^i = 0
\end{align}
\end{subequations}
\textbf{Dynamics:}
\begin{equation}
    {\bm{x}}_{k+1}^i = {\bar {\bm{A}}}_k^i {\bm{x}}_k^i + {{\bar {\bm{B}}}}_k^i {\bm{u}}_k^i + {{\bar {\bm{C}}}}_k^i {\bm{u}}_{k+1}^i + \bar \Sigma_k^i \sigma^i + {\bar {\bm{z}}}_k^i + \nu_k^i
    \label{eq:dynamics1}
\end{equation}
\textbf{State Constraints:}
\begin{subequations}
\label{eq:state_constraints2}
\begin{align}
    & m_{\text{dry}} \leq m_k^i
\end{align}
\end{subequations}
\textbf{Control Constraints:}
\begin{subequations}
\label{eq:control_constraints2}
\begin{align}
    & T_{\text{min}} \leq B_{\bm{g}}(\tau_k) {\bm{u}}_k^i \\
    & \| {\bm{u}}_k^i \|_2 \leq T_{\text{max}} \\
    & \cos \delta_{\text{max}} \| {\bm{u}}_k^i \|_2 \leq {\bm{e}}_1 \cdot {\bm{u}}_k^i
\end{align}
\end{subequations}
\textbf{Trust Regions:}
\begin{subequations}
\label{eq:trust_regions}
\begin{align}
    & \delta {\bm{x}}_k^i \cdot \delta {\bm{x}}_k^i + \delta {\bm{u}}_k^i \cdot \delta {\bm{u}}_k^i \leq \Delta_k^i \\
    & \|\delta \sigma^i \|_1 \leq \Delta_\sigma^i.
\end{align}
\end{subequations}

In the cost function, the term \( \sigma^i \) represents the time-scaling factor, which is minimized alongside the control inputs \( \bm{u}_k^i \). The term \( \bm{\omega}_\nu \| \bm{\bar{\nu}}^i \|_1 \) penalizes the introduction of virtual control \( \bm{\nu}_k^i \), which is used to avoid infeasibility in the convexification process. The terms \( \bm{\omega}_{\bm{\Delta}}^i \| \bm{\bar{\Delta}}^i \|_2 \) and \( \bm{\omega}_{\bm{\Delta}_{\sigma}} \| \bm{\Delta}_\sigma \|_1 \) penalize deviations from the previous iterates for the state and time-scaling factors. The boundary conditions ensure that the initial wet mass \( m_0^i = m_{\text{wet}} \), initial position \( \bm{r}_{{\mathcal{I}}, 0}^i = \bm{r}_{{\mathcal{I}}, i} \), initial velocity \( \bm{v}_{{\mathcal{I}}, 0}^i = \bm{v}_{{\mathcal{I}}, i} \), and initial angular velocity \( \bm{\omega}_{{\mathcal{B}}, 0}^i = \bm{\omega}_{{\mathcal{B}}, i} \) are satisfied. At the final time step \( K \), the position \( \bm{r}_{{\mathcal{I}}, K}^i = 0 \), the velocity \( \bm{v}_{{\mathcal{I}}, K}^i = \bm{v}_{{\mathcal{I}}, f} \), the quaternion \( \bm{q}_{{\mathcal{B}} \leftarrow {\mathcal{I}}, K}^i = \bm{q}_{{\mathcal{B}} \leftarrow {\mathcal{I}}, f} \), and the angular velocity \( \bm{\omega}_{{\mathcal{B}}, K}^i = 0 \) are enforced, along with constraints that the thrust vector components along the body frame axes \( \bm{e}_2 \) and \( \bm{e}_3 \) are zero.

The dynamics of the system are defined by the discrete-time state equation \( \bm{x}_{k+1}^i = \bar{\bm{A}}_k^i \bm{x}_k^i + \bar{\bm{B}}_k^i \bm{u}_k^i + \bar{\bm{C}}_k^i \bm{u}_{k+1}^i + \bar{\Sigma}_k^i \sigma^i + \bar{\bm{z}}_k^i + \nu_k^i \), where \( \bar{\bm{A}}_k^i \), \( \bar{\bm{B}}_k^i \), and \( \bar{\bm{C}}_k^i \) are the system matrices, \( \bar{\bm{z}}_k^i \) is a disturbance term, and \( \nu_k^i \) is the virtual control used to maintain feasibility. The state constraints ensure the mass is greater than or equal to the dry mass \( m_{\text{dry}} \).

Control constraints ensure that the thrust vector magnitude \( \bm{u}_k^i \) lies between \( T_{\text{min}} \) and \( T_{\text{max}} \), and the thrust direction is constrained by \( \cos \delta_{\text{max}} \| \bm{u}_k^i \|_2 \leq \bm{e}_1 \cdot \bm{u}_k^i \), where \( \delta_{\text{max}} \) defines the maximum allowable angle between the thrust and the body frame axis. Trust regions are defined by bounds on the change in state \( \delta \bm{x}_k^i \) and control \( \delta \bm{u}_k^i \) to ensure the convex sub-problems remain bounded and feasible throughout the iteration process. The constraints \( \delta \bm{x}_k^i \cdot \delta \bm{x}_k^i + \delta \bm{u}_k^i \cdot \delta \bm{u}_k^i \leq \Delta_k^i \) and \( \| \delta \sigma^i \|_1 \leq \Delta_\sigma^i \) enforce the trust regions around the previous iterate. For an extended discussion and derivation of the 6 DoF powered descent guidance problem and SCvx see \cite{szmuk2018successivefinaltime}.

Outputs from solving Equations \ref{eq:cost_function}-\ref{eq:trust_regions} are arranged in a 2D matrix for training TrajDiffuser, Equation \ref{eq: trajectory}.

\begin{equation}
    {\bm{z}}_{0:H} = \begin{bmatrix}
    {\bm{r}}_0 & {\bm{r}}_1 & \cdots & {\bm{r}}_H \\
    {\bm{v}}_0 & {\bm{v}}_1 & \cdots & {\bm{v}}_H \\
    {\bm{q}}_0 & {\bm{q}}_1 & \cdots & {\bm{q}}_H \\
    {\bm{\omega}}_0 & {\bm{\omega}}_1 & \cdots & {\bm{\omega}}_H \\
    m_0 & m_1 & \cdots & m_H \\
    {\bm{T}}_0 & {\bm{T}}_1 & \cdots & {\bm{T}}_H
    \end{bmatrix}
    \label{eq: trajectory}
\end{equation}

Storing the states and control inputs ${\bm{z}}$ over the horizon time $H$. Combining these matrices into a set of the required batch size, we have the diffusion state of the original trajectory $x_0 = [{\bm{z}}^0_{0:H}, \; {\bm{z}}^1_{0:H}, \; \dots, {\bm{z}}^b_{0:H}]$, for batch size $b$.

\section{State-Triggered Constraints}
\label{sec:constraints}

The state-triggered constraint formulation includes a \textit{trigger condition} which, when satisfied, leads to enforcing a \textit{constraint condition}. Logically, state-triggered constraints for inequality-based conditions are expressed as
\begin{equation}
    (g({\bm{z}}) \leq 0) \rightarrow (c({\bm{z}}) \leq 0),
    \label{eq: state triggered}
\end{equation}
where ${\bm{z}} \in {\mathbb{R}}^{n_z}$ are the solution variables, $g(\cdot) : {\mathbb{R}}^{n_z} \rightarrow {\mathbb{R}}$ is the trigger condition, and $c(\cdot) : {\mathbb{R}}^{n_z} \rightarrow {\mathbb{R}}$ is the constraint condition.

Previous work has extended the state-triggered constraint formulation to include a slack variable, enabling continuous state-triggered constraint (cSTCs) in numerical optimization frameworks \cite{Szmuk2019}. The converted constraint is written as follows:
\begin{equation}
    h({\bm{z}}) := - \min (g({\bm{z}}), 0) \cdot c({\bm{z}}) \leq 0.
    \label{eq: cSTC}
\end{equation}
In this work, we use the constraint formulation in Equation \eqref{eq: cSTC} in the numerical optimization problem but use an alternative formulation for diffusion model composition; the logic in Equation \eqref{eq: state triggered} can be reformulated to return True if the full state-triggered constraint is violated:
\begin{equation}
    (g({\bm{z}}) \leq 0) \cap 
\neg (c({\bm{z}}) \leq 0).
    \label{eq: state triggered diffusion}
\end{equation}
\subsection{Velocity-Triggered Angle of Attack Constraint}

The velocity-triggered angle of attack constraint introduces aerodynamic load alleviation by limiting a spacecraft's angle of attack when the dynamic pressure is high. To prevent an overly conservative solution, the angle of attack is not limited when dynamic pressure is negligible. The state-triggered constraint is formulated as follows,
\begin{align}
    \|{\bm{v}}_{\mathcal{I}} (t) \|_2 > V_\alpha &\rightarrow -{\bm{e}}_1 \cdot C_{{\mathcal{B} \leftarrow {\mathcal{I}}}} ({\bm{q}}_{{\mathcal{B} \leftarrow {\mathcal{I}}}}) {\bm{v}}_{\mathcal{I}} (t) \\
    &\geq \cos \alpha_{\max} \| {\bm{v}}_{\mathcal{I}} (t) \|_2,
\end{align}
where ${\bm{V}}_{\alpha} \in {\mathbb{R}}_{++}$ is the speed above which the angle of attack is limited to $\alpha (t) \in [0, \alpha_{\max}]$. As derived in Szmuk et al., the cSTC formulation is the following inequality constraint \cite{Szmuk2019}.

\begin{align}
h_{\alpha}({\bm{v}}_{\mathcal{I}}(t), {\bm{q}}_{{\mathcal{B} \leftarrow {\mathcal{I}}}}(t)) &:= -\min\left(g_{\alpha}({\bm{v}}_{\mathcal{I}}(t)), 0 \right) \cdot \\
&c_{\alpha}({\bm{v}}_{\mathcal{I}}(t), {\bm{q}}_{{\mathcal{B} \leftarrow {\mathcal{I}}}}) \leq 0,
\end{align}

where
\begin{align}
g_{\alpha}({\bm{v}}_{\mathcal{I}}(t)) &:= {\bm{V}}_{\alpha} - \|{\bm{v}}_{\mathcal{I}}(t)\|_2, \\
c_{\alpha}({\bm{v}}_{\mathcal{I}}(t), {\bm{q}}_{{\mathcal{B} \leftarrow {\mathcal{I}}}}) &:= \cos \alpha_{\max} \|({\bm{v}}_{\mathcal{I}}(t)\|_2 \\
&+ {\bm{e}}_1 \cdot C_{{\mathcal{B} \leftarrow {\mathcal{I}}}} ({\bm{q}}_{{\mathcal{B} \leftarrow {\mathcal{I}}}}) {\bm{v}}_{\mathcal{I}}(t).
\end{align}
Using Equation \eqref{eq: state triggered diffusion}, we can formulate violations of the velocity-triggered angle of attack constraint as:
\begin{align}
    (\|{\bm{v}}_{\mathcal{I}} (t) \|_2 > V_\alpha) \cap& \neg (-{\bm{e}}_1 \cdot C_{{\mathcal{B} \leftarrow {\mathcal{I}}}} ({\bm{q}}_{{\mathcal{B} \leftarrow {\mathcal{I}}}}) {\bm{v}}_{\mathcal{I}} (t) \\
    &\geq \cos \alpha_{\max} \| {\bm{v}}_{\mathcal{I}} (t) \|_2).
    \label{eq: velocity triggered diffusion}
\end{align}
\section{Experiments}
\label{sec:experiments}

The following experiments detail the training and benchmarking of a minimum-time 6 DoF powered descent guidance diffusion model, TrajDiffuser. They also benchmark sampling from this model and develop a series of compositional models, or C-TrajDiffusers, enabling the generalization of the trained trajectory model for out-of-distribution test cases. The code and models used in this work will be released following publication.

\subsection{Minimum-Time Six Degree-of-Freedom Powered Descent Guidance Diffusion Model}\label{sec:TrajDiffuser}

The training dataset for the trajectory diffusion model is generated in Python using CVXpy and ECOS \cite{diamond2016cvxpy, agrawal2018rewriting, domahidi2013ecos}. The SCvx algorithm is called iteratively using Equations \eqref{eq:cost_function}-\eqref{eq:trust_regions}, using the ECOS solver to solve each SOCP subproblem \cite{Szmuk2019}. All variables and parameters use non-dimensional quantities, and the SCvx parameters are equivalent to the parameters in Szmuk and A\c{c}{\i}kme\c{s}e, unless otherwise specified \cite{szmuk2018successivefinaltime}. Table \ref{tab:algorithm_settings} defines the powered descent guidance problem parameters, including $K=20$ time discretization nodes.

\begin{table}
        \caption{\label{tab:algorithm_settings} Problem Parameters}
        \centering
        \begin{tabular}{ll}
        \hline
            \hline \\
            \textbf{Parameter} & \textbf{Value} \\
            \hline \\
            Gravity ($g$) & $- {\bm{e}_1}$ \\
            Flight Time Guess ($t_{f \text{guess}}$) & 3 \\
            Fuel Consumption Rate (${\alpha_m}$) & 0.01 \\
            Thrust to COM Vector (${r_{T, {\mathcal{B}}}}$) & -0.01 ${\bm{e}_1}$ \\
            Angular Moment of Inertia ($J_{\mathcal{B}}$) & 0.01 $I_{3\times 3}$ \\
            Number of discretization nodes ($K$) & 20 \\
            \hline
            \hline
        \end{tabular}
    \end{table}

To generate a training dataset with a range of initial positions, velocities, orientations, angular velocities, initial mass, and control input ranges, uniform sampling was performed for each sample according to the distributions in Table \ref{tab:sampled}.

    \begin{table}
        \caption{\label{tab:sampled} Sampled Dataset}
        \centering
         \resizebox{\columnwidth}{!}{
        \begin{tabular}{ll}
            \hline
            \hline \\
            \textbf{Parameter} & \textbf{Sampled Trajectory Distribution} \\
            \hline \\
            Initial Z Position ($r_{z,0}$) & $\sim {\mathcal{U}}$ [1, 4] \\
            Initial X Position ($r_{x,0}$) & $\sim {\mathcal{U}}$ [-2, 2] \\
            Initial Y Position ($r_{y,0}$) & $\sim {\mathcal{U}}$ [-2, 2] \\
            Initial Z Velocity ($v_{z,0}$) & $\sim {\mathcal{U}}$ [-1, -0.5] \\
            Initial X Velocity ($v_{x,0}$) & $\sim {\mathcal{U}}$ [-0.5, -0.2] \\
            Initial Y Velocity ($v_{y,0}$) & $\sim {\mathcal{U}}$ [-0.5, -0.2] \\
            Initial Quaternion (${{\bm{q}}_{{\mathcal{B}},0}}$) & ${\sim}$ euler to q (0,  ${\mathcal{U}}$ [-30, 30], \\ & ${\mathcal{U}}$ [-30, 30]) \\
            Initial Z Angular Velocity ($\omega_{z,0}$) & 0 \\
            Initial X Angular Velocity ($\omega_{x,0}$) & $\sim {\mathcal{U}}$ [-20, 20] \\
            Initial Y Angular Velocity ($\omega_{y,0}$) & $\sim {\mathcal{U}}$ [-20, 20] \\
            Wet Mass ($m_{\text{wet}}$) & $\sim {\mathcal{U}}$ [2, 5] \\
            Dry Mass ($m_{\text{dry}}$) & $\sim {\mathcal{U}}$ [0.1, 2] \\
            Max Gimbal Angle (${\delta_{\max}}$) & $\sim {\mathcal{U}}$ [10, 90] \\
            Max Thrust (${T_{\max}}$) & $\sim {\mathcal{U}}$ [3, 10] \\
            Min Thrust (${T_{\min}}$) & $\sim {\mathcal{U}}$ [0.01, 1] \\
            \hline
            \hline
        \end{tabular}
        }
    \end{table}

\subsubsection{Diffusion Model Architecture}

A timestep embedding is used to encode the current diffusion timestep $t$ into an input for the diffusion model. The timestep embedding for a given timestep $t$, the embedding $e_t$ is defined as:
\[
e_t = \text{Concat}(\sin(\omega \cdot t), \cos(\omega \cdot t)),
\]
where $\omega$ is a vector of frequencies:
\[
\omega_i = \frac{1}{10000^{\frac{2i}{d}}} \quad \text{for} \quad i = 0, 1, \dots, \frac{d}{2} - 1,
\]
where $d$ is the embedding dimension. The neural network defining the parameterized energy function $E_{\bm{\theta}} ({\bm{x}},t)$ is a U-Net due to their exceptional performance in trajectory planning tasks, including the advantage of variable planning horizon lengths \cite{janner2022planning, kaiser2020model}. Table \ref{tab: Unet} shows the U-Net architecture for the parameterized energy function. The convolutional blocks also include transpose, layer norm, and linear layers, which are not included in the table for brevity.

\begin{table}[h!]
\centering
\caption{U-Net Diffusion Model Architecture} \label{tab: Unet}
\begin{tabular}{ l l }
\hline
\hline \\
\textbf{Layer}                  & \textbf{Weight Shape}        \\
\hline \\
\textbf{Input Layer}            & \texttt{(None, H, W, C)}     \\
\textbf{conv2\_d}                    & \texttt{(3, 3, 32, 1)}       \\
%\textbf{conv2\_d\_transpose}         & \texttt{(4, 4, 256, 256)}    \\
%\textbf{conv2\_d\_transpose 1}       & \texttt{(4, 4, 128, 256)}    \\
%\textbf{conv2\_d\_transpose 2}       & \texttt{(4, 4, 64, 128)}     \\
%\textbf{conv2\_d\_transpose 3}       & \texttt{(4, 4, 32, 64)}      \\
\textbf{conv\_block 1/conv2\_d}      & \texttt{(3, 3, 1, 32)}       \\
\textbf{conv\_block 1/conv2\_d 1}    & \texttt{(3, 3, 32, 32)}      \\
%\textbf{conv\_block 1/layer\_norm}   & \texttt{(32)}                \\
%\textbf{conv\_block 1/linear}        & \texttt{(32, 32)}            \\
\textbf{conv\_block 2/conv2\_d}      & \texttt{(3, 3, 16, 64)}      \\
\textbf{conv\_block 2/conv2\_d 1}    & \texttt{(3, 3, 64, 64)}      \\
%\textbf{conv\_block 2/layer\_norm}   & \texttt{(64)}                \\
%\textbf{conv\_block 2/linear}        & \texttt{(32, 64)}            \\
\textbf{conv\_block 3/conv2\_d}      & \texttt{(3, 3, 32, 128)}     \\
\textbf{conv\_block 3/conv2\_d 1}    & \texttt{(3, 3, 128, 128)}    \\
%\textbf{conv\_block 3/layer\_norm}   & \texttt{(128)}               \\
%\textbf{conv\_block 3/linear}        & \texttt{(32, 128)}           \\
\textbf{conv\_block 4/conv2\_d}      & \texttt{(3, 3, 64, 256)}     \\
\textbf{conv\_block 4/conv2\_d 1}    & \texttt{(3, 3, 256, 256)}    \\
%\textbf{conv\_block 4/layer\_norm}   & \texttt{(256)}               \\
%\textbf{conv\_block 4/linear}        & \texttt{(32, 256)}           \\
\textbf{conv\_block 5/conv2\_d}      & \texttt{(3, 3, 512, 256)}    \\
\textbf{conv\_block 5/conv2\_d 1}    & \texttt{(3, 3, 256, 256)}    \\
%\textbf{conv\_block 5/layer\_norm}   & \texttt{(256)}               \\
%\textbf{conv\_block 5/linear}        & \texttt{(32, 256)}           \\
\textbf{conv\_block 6/conv2\_d}      & \texttt{(3, 3, 256, 128)}    \\
\textbf{conv\_block 6/conv2\_d 1}    & \texttt{(3, 3, 128, 128)}    \\
%\textbf{conv\_block 6/layer\_norm}   & \texttt{(128)}               \\
%\textbf{conv\_block 6/linear}        & \texttt{(32, 128)}           \\
\textbf{conv\_block 7/conv2\_d}      & \texttt{(3, 3, 128, 64)}     \\
\textbf{conv\_block 7/conv2\_d 1}    & \texttt{(3, 3, 64, 64)}      \\
%\textbf{conv\_block 7/layer\_norm}   & \texttt{(64)}                \\
%\textbf{conv\_block 7/linear}        & \texttt{(32, 64)}            \\
\textbf{conv\_block 8/conv2\_d}      & \texttt{(3, 3, 64, 32)}      \\
\textbf{conv\_block 8/conv2\_d 1}    & \texttt{(3, 3, 32, 32)}      \\
%\textbf{conv\_block 8/layer\_norm}   & \texttt{(32)}                \\
%\textbf{conv\_block 8/linear}        & \texttt{(32, 32)}            \\
\textbf{Output Layer}                & \texttt{(3, 3, 32, C)}       \\               \\ \hline \hline
\end{tabular}
\end{table}

The model includes a downsampling path and an upsampling path. The downsampling path reduces the input's spatial resolution and increases the number of feature channels. The upsampling path restores the original resolution while progressively combining high-level features from the path. Downsampling includes convolutional layers with layer normalization and Swish activations, followed by a max-pooling layer. Once the bottleneck is reached, transported convolutional layers are used for upsampling, and skip connections from the downsampling path are concatenated with the upsampled features at each level.

Cosine beta scheduling ensures that \(\alpha_{\text{cumprod}}(t)\) starts close to 1 (at \(t = 0\)) and gradually decreases to 0 as \(t \to T\), adding noise over the forward process. For the diffusion process with \(T\) timesteps, the cosine beta schedule is defined by first computing the cumulative products \(\alpha_{\text{cumprod}}(t)\), representing how much of the original data remains after timestep \(t\). The function for \(\alpha_{\text{cumprod}}(t)\) is based on a scaled cosine function and is given by:
\[
\alpha_{\text{cumprod}}(t) = \cos^2\left(\frac{\left(\frac{t}{T} + s\right)}{1 + s} \cdot \frac{\pi}{2}\right),
\]
where $s$ is a small constant set to 0.008 in this work. The noise variance \(\beta_t\) for each timestep is computed from the changes in \(\alpha_{\text{cumprod}}(t)\):
\[
\beta_t = 1 - \frac{\alpha_{\text{cumprod}}(t+1)}{\alpha_{\text{cumprod}}(t)}.
\]
For additional information on cosine beta scheduling, see Nichol and Dhariwal \cite{nichol2021improved}.

\subsubsection{TrajDiffuser: Trajectory Diffusion Model}

This section details the trained minimum-time 6 DoF powered descent guidance diffusion model, TrajDiffuser. For training and model outputs, values are scaled according to sklearn.preprocessing.RobustScaler fit on the training data \cite{scikit-learn}. RobustScaler was used instead of StandardScaler since $x$ values outside of the range [-1, 1] can significantly inhibit the learning process, and the samples obtained from the numerical optimizer often have outliers. The training process used batch size 50 and was terminated when TrajDiffuser reverse diffusion samples obtained losses less than 0.2, which required less than 20,000 samples. The complete model generates trajectories of batch size x 20 timesteps x 17 states.

\begin{figure}[htbp]
    \centering
    \includegraphics[width=0.48\textwidth]{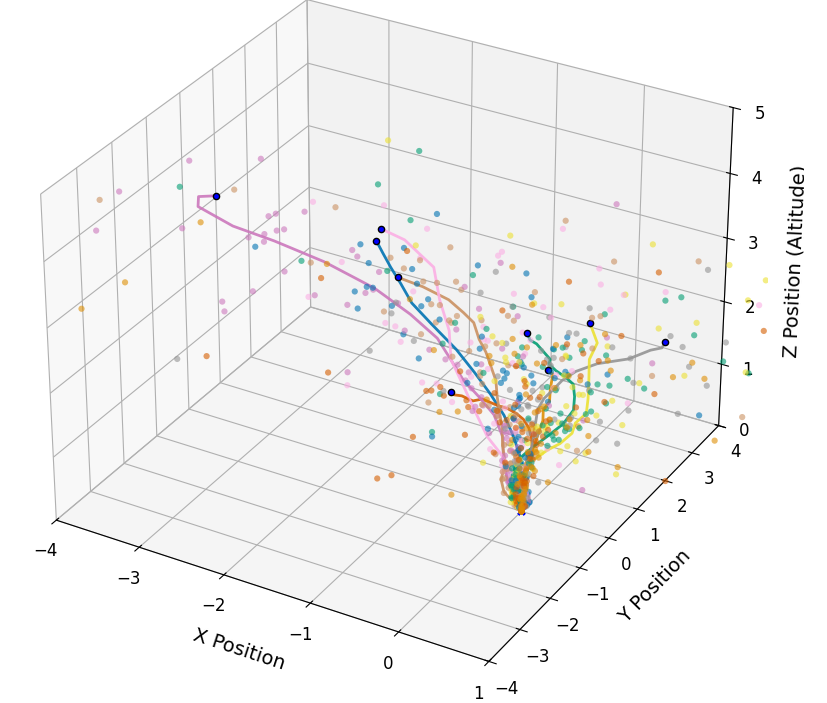}
    \caption{Diffusion model reverse sampled trajectories from TrajDiffuser. The scattered points show the trajectory for every 200 steps in the backward process, increasing opacity as the trajectory converges. Each color indicates a separate trajectory.}
    \label{fig:diffusion process}
\end{figure}

A visualization of reverse process sampling from TrajDiffuser is shown in Figure \ref{fig:diffusion process}. As the diffusion timestep reduces from 1000 to zero, the distribution converges to a distribution of smooth trajectory samples. Each set of trajectories $x_t$ is unscaled using the RobustScaler before plotting. Starting at $t=1000$, a scatter of samples $x_t$ are plotted, where ${x_{T}} \sim {\mathcal{N}}(0, I)$. For every 200 samples, $x_t$ is plotted again, with an increasing opacity. Finally, the trajectories generated for ${x_0}$ are plotted as lines. All trajectories reach the final landing location, and even starting locations that are fairly close together result in trajectories that exhibit multi-modal behaviors.

% Figure for Generated Trajectories - Position

\begin{figure*}[htbp]
    \centering
    \includegraphics[width=0.8\textwidth]{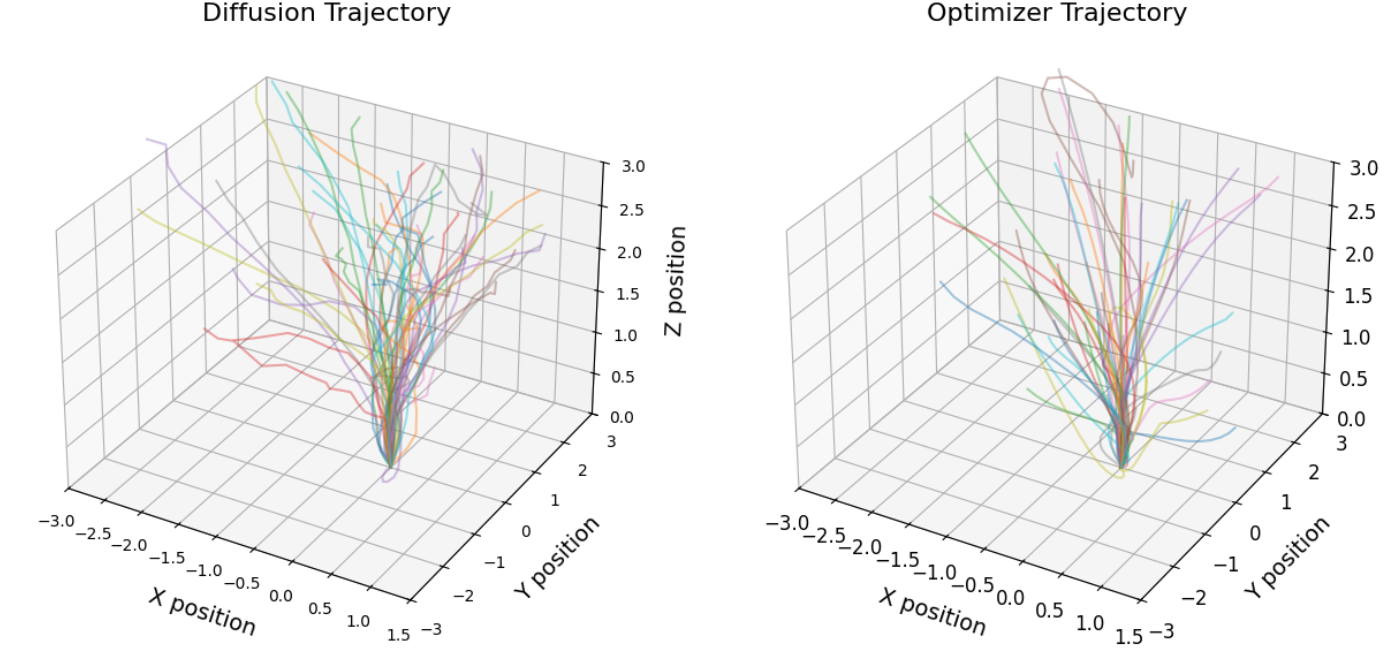}
    \caption{Generated Trajectories - Position. The lefthand plot shows 50 trajectories generated by the diffusion model, and the righthand plot shows 50 trajectories uniformly sampled by the trajectory optimization framework.}
    \label{fig:generated_trajectories_position}
\end{figure*}

\begin{figure*}[htbp]
    \centering
    \includegraphics[width=0.8\textwidth]{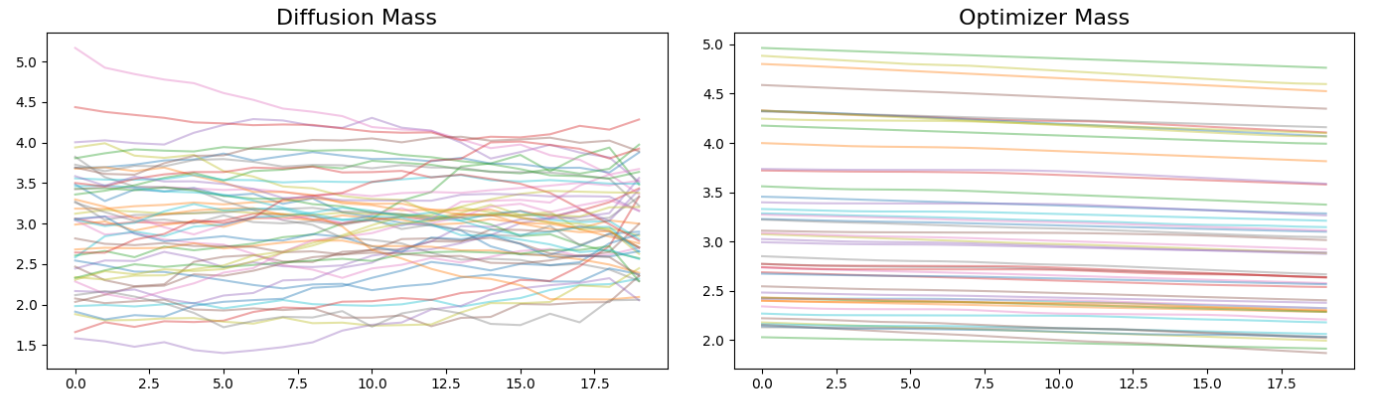}
    \caption{Mass Over Time. The lefthand plot shows 50 mass values over time generated by the diffusion model, and the righthand plot shows 50 mass values over time uniformly sampled by the trajectory optimization framework.}
    \label{fig:mass_over_time}
\end{figure*}

\begin{figure*}[htbp]
    \centering
    \includegraphics[width=0.8\textwidth]{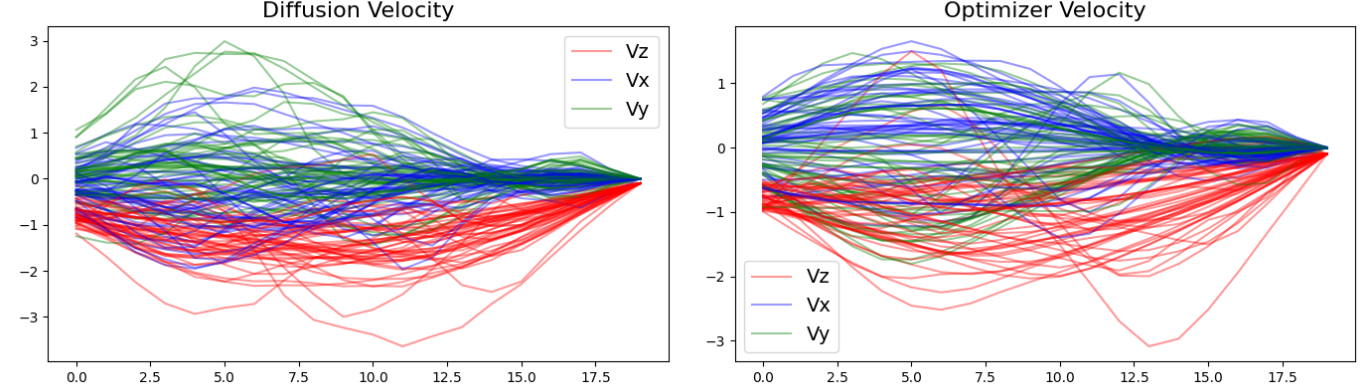}
    \caption{Velocity Components Over Time. The lefthand plot shows 50 velocity values over time generated by the diffusion model, and the righthand plot shows 50 velocity values over time uniformly sampled by the trajectory optimization framework.}
    \label{fig:velocity_components_over_time}
\end{figure*}

\begin{figure*}[htbp]
    \centering
    \includegraphics[width=0.8\textwidth]{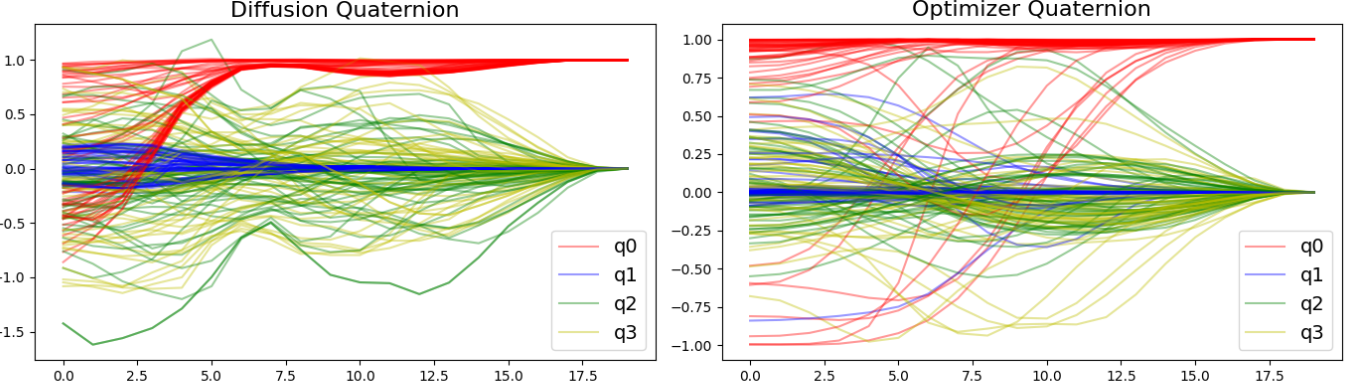}
    \caption{Quaternion Components Over Time. The lefthand plot shows 50 quaternion values over time generated by the diffusion model, and the righthand plot shows 50 quaternion values over time uniformly sampled by the trajectory optimization framework.}
    \label{fig:quaternion_components_over_time}
\end{figure*}

\begin{figure*}[htbp]
    \centering
    \includegraphics[width=0.8\textwidth]{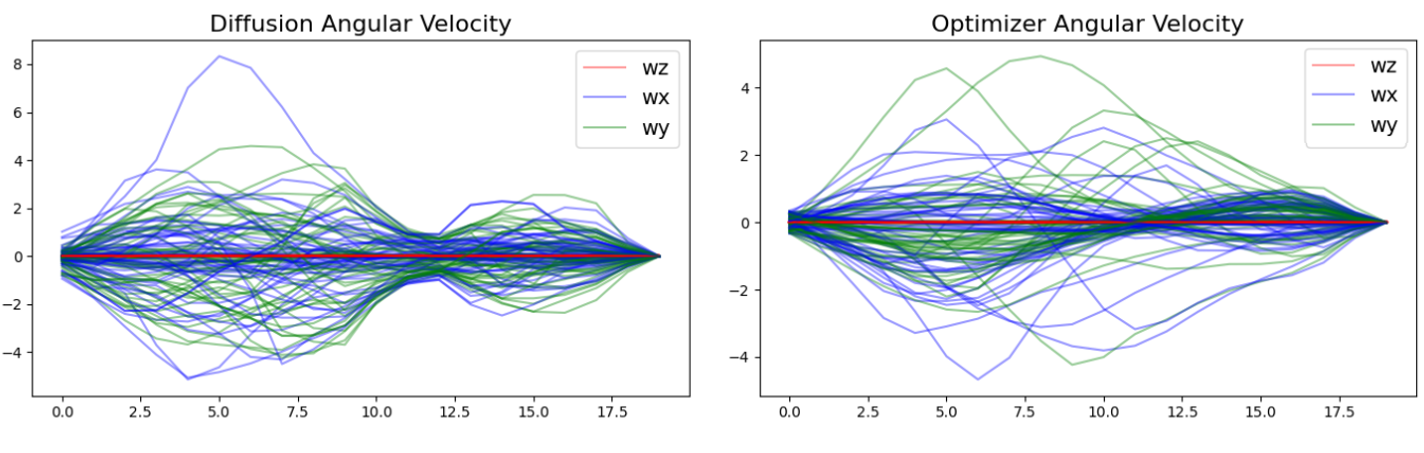}
    \caption{Angular Velocity Components Over Time. The lefthand plot shows 50 angular velocity values over time generated by the diffusion model, and the righthand plot shows 50 angular velocity values over time uniformly sampled by the trajectory optimization framework.}
    \label{fig:angular_velocity_components_over_time}
\end{figure*}

\begin{figure*}[htbp]
    \centering
    \includegraphics[width=.8\textwidth]{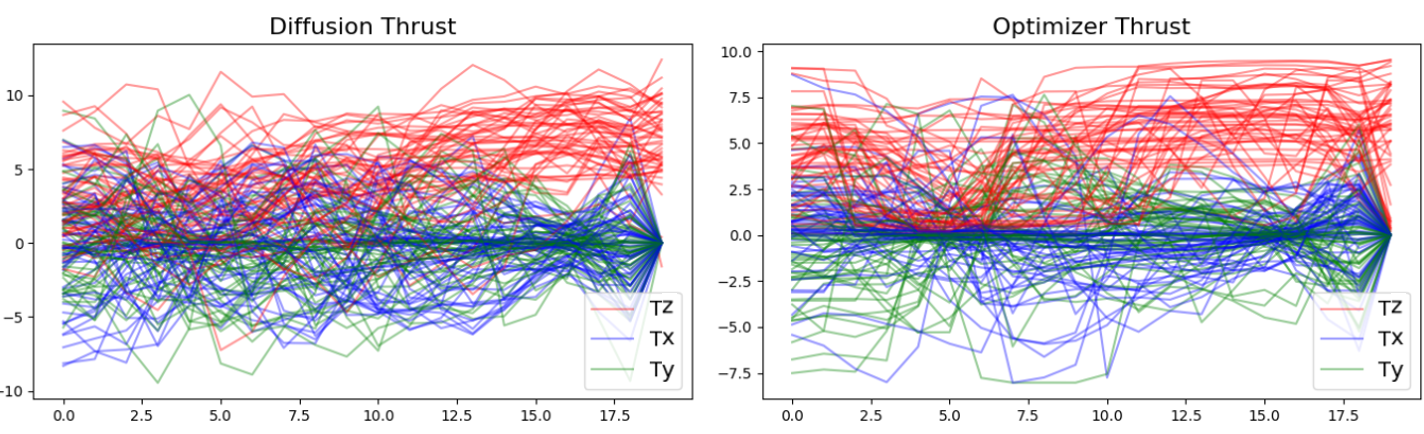}
    \caption{Thrust Components Over Time. The lefthand plot shows 50 thrust values over time generated by the diffusion model, and the righthand plot shows 50 thrust values over time uniformly sampled by the trajectory optimization framework.}
    \label{fig:thrust_components_over_time}
\end{figure*}

To evaluate the performance of the diffusion model, the learned trajectory representation by TrajDiffuser is compared against samples returned by the numerical optimizer. Overall, Figures \ref{fig:generated_trajectories_position}-\ref{fig:thrust_components_over_time} show samples from the learned distribution of states and control inputs obtained by the diffusion model.

Figure \ref{fig:generated_trajectories_position} shows the 50 trajectories generated by TrajDiffuser, which lie in roughly the same range as numerical optimizer-generated trajectories. Some trajectory samples from the diffusion model include more noise, indicating that further model training or smoothing could improve the trajectories. Another observation is that samples from the numerical optimizer include more low-altitude samples on the positive x-axis. These samples are likely more out of distribution for the diffusion model. 

% Figure for Generated Trajectories - Position

The diffusion-sampled mass vs. numerical optimizer-computed mass share similar scales in Figure \ref{fig:mass_over_time}. Still, the rate of change of the mass for TrajDiffuser frequently has a higher magnitude than the almost flat lines for the numerical optimizer's generated mass. If the diffusion model predicts an increasing mass, the practitioner must correct this value to ensure it is less than or equivalent to the previous timestep mass value to respect the dynamics constraints.

% Figure for Mass Over Time

The TrajDiffuser and numerical optimizer sampled velocities over time are shown in Figure \ref{fig:velocity_components_over_time}. A similar policy of negative velocities in the z direction and closer to positive velocities in the x and y directions is demonstrated across both sets of samples. Additionally, a change between high x velocities and y velocities close to time step 15 was well-captured by the diffusion model. One main difference is the diffusion model's larger positive x and y velocities. No velocity constraint was imposed on the optimizer, so these larger velocities would not violate any velocity constraints for the current model.

% Figure for Velocity Components Over Time

Figure \ref{fig:quaternion_components_over_time} shows samples for $q_{0}, q_{1}, q_{2},$ and $q_{3}$ over time. While one erroneous diffusion sample for $q_{2}$ goes outside of the [-1, 1] range allowable for quaternions, all other samples match the numerical optimizer's policy well. Another observation is that the diffusion model moves the spacecraft to the vertical position required for landing about 2.5 timesteps before the numerical optimizer does. An overview of the magnitudes of these errors is provided later in this section.

% Figure for Quaternion Components Over Time

Diffusion model generated vs. numerical optimizer sampled angular velocities are shown in Figure \ref{fig:angular_velocity_components_over_time}. Except for the one outlier x angular velocity value, both sets of samples lie in the [-4, 4] range. Further, the switching between positive and negative x and y values, as observed for the velocity as well, is captured in the diffusion model. wz remains zero since the sampled model assumed no angular velocity in the z-direction. If angular velocity dynamics in this direction are needed, a mixture composition diffusion model could introduce those dynamics.

% Figure for Angular Velocity Components Over Time

Lastly, the control inputs, thrust for powered descent guidance, for TrajDiffuser and numerical optimization are included in Figure \ref{fig:thrust_components_over_time}. The optimizer's policy of positive z thrust and alternating x and y thrust is mirrored in the diffusion samples.

% Figure for Thrust Components Over Time

\begin{figure*}
    \centering
    \includegraphics[width=.9\linewidth]{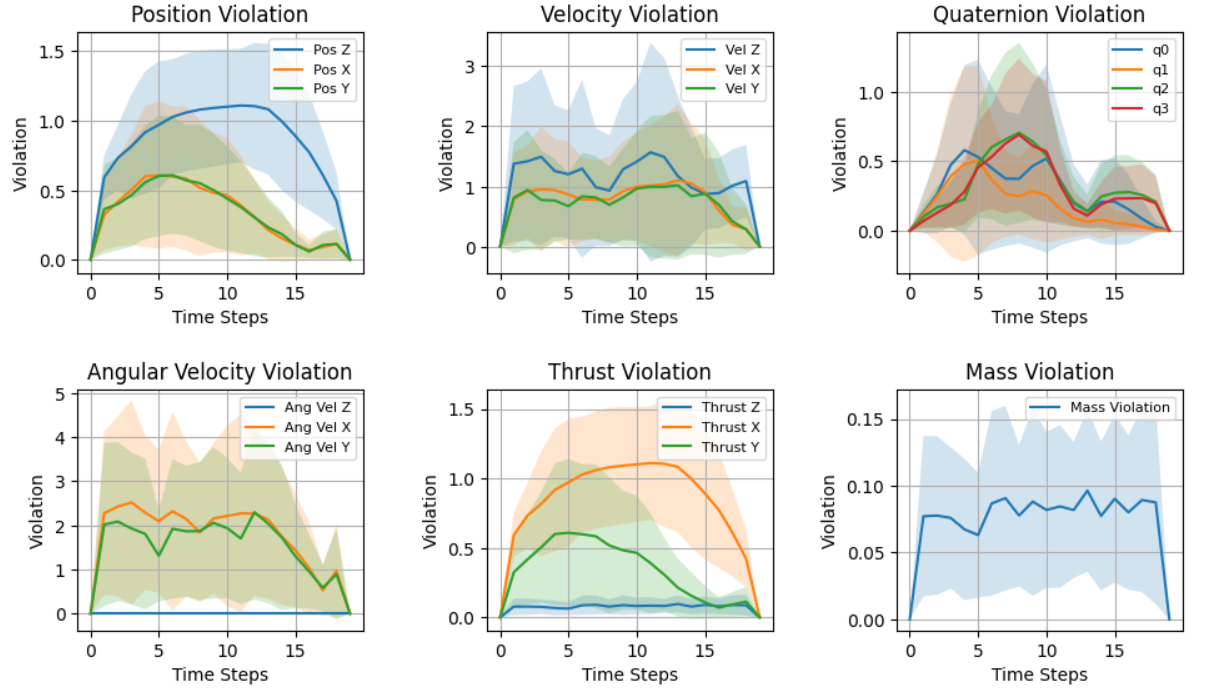}
    \caption{Dynamics constraint violation means and standard deviations for the diffusion trajectory model for 50 sampled trajectories.}
    \label{fig:constraint_violation}
\end{figure*}

To evaluate the degree of dynamics constraint violation at every step, the model dynamics were propagated for the given control input at each timestep and compared with the TrajDiffuser trajectories for 50 samples. The magnitudes of the difference between the propagated states and control inputs $x_{t+1}$ and the diffusion model's states and control inputs ${\hat{x}_{t+1}}$, $| x_{t+1} - {\hat{x}_{t+1}} |$ were recorded for each time step $t \in H$, where the time horizon $H = 20$. The means and standard deviation for the degree of violation for each decision variable are plotted for each state and control input in Figure \ref{fig:constraint_violation}.

The z direction, or altitude, has the most significant errors for position, velocity, and mass. Angular velocity and thrust violation are highest for the x direction. Overall, each state and control variable has the highest degree of constraint violation towards the middle of the trajectory or halfway through horizon $H$. Note that the standard deviation is very large for each variable, and each timestep is a local measure of constraint violation, only evaluating constraint satisfaction on the next step of dynamics propagation given the specific set of control inputs.

Now that constraint satisfaction has been evaluated for TrajDiffuser, we desire to determine the degree of optimality achieved by the diffusion model. Since SCvx solves a non-convex optimization problem, no guarantees exist for convergence to a globally optimal solution. To determine the degree of optimality, the proximity to the nearest locally optimal solution is quantified as a set of mean-squared-errors (MSEs) and standard deviations of the error between each sampled diffusion trajectory and the locally optimal solution obtained by warm-starting the trajectory optimizer with that guess. The warm-started optimizer used the widest bounds to define the minimum mass, thrust limit, and maximum gimbal angle state and control constraints. Table \ref{tab: MSE} shows each variable's MSE and standard deviation over a batch size of 50 trajectories and over the 20 timesteps in the horizon $H$.

\begin{table}[ht]
\centering
\caption{Diffusion Degree of Optimality} \label{tab: MSE}
\begin{tabular}{c c c}
\hline \hline\\
\textbf{Variable} & \textbf{Mean Squared Error} & \textbf{Standard Deviation} \\ 
\hline \\
M  & 0.6136  & 0.7814  \\
Z  & 4.502   & 2.039   \\
X  & 0.4740  & 0.6798  \\
Y  & 0.6785  & 0.8228  \\
Vz  & 0.9905  & 0.9750  \\
Vx  & 0.5419  & 0.7350  \\
Vy  & 0.6484  & 0.8034  \\
q0  & 0.5991  & 0.6047  \\
q1  & 0.05728 & 0.2357  \\
q2 & 0.3034  & 0.4236  \\
q3 & 0.2461  & 0.4708  \\
wz & 0 & 0 \\
wx & 3.557   & 1.413   \\
wy & 2.254   & 1.471   \\
Tz & 11.10   & 3.331   \\
Tx & 9.556   & 3.084   \\
Ty & 13.69   & 3.695   \\
\hline \hline
\end{tabular}

\end{table}

The highest MSEs occur for the z position, or altitude, and the thrust components. The thrust components are expected to have the highest magnitude MSE since they have the largest sampling range for the model. Also, since the widest bounds were used for the state and control constraints, the z-direction prediction position may be a sample that better represents a more constrained constraint profile. Overall, most MSEs are less than 1 with relatively small standard deviations, indicating close to locally optimal sampling performance.

\subsection{Computational Complexity: Reverse Diffusion Sampling vs. Numerical Optimization}

\begin{figure}
    \centering
    \includegraphics[width=\linewidth]{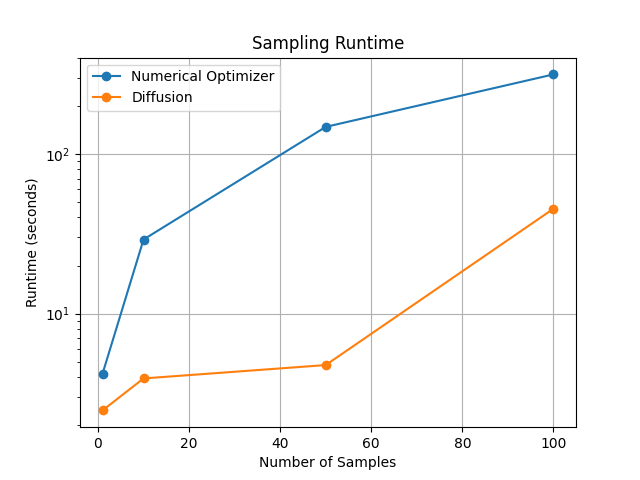}
    \caption{The runtimes of sampling from the numerical optimization problem vs. the trained diffusion model for 1, 10, 50, and 100 samples.}
    \label{fig:samplingruntime}
\end{figure}

To evaluate the computational efficiency of trajDiffuser, the runtimes for sampling from the diffusion model vs. the numerical optimizer are included in Figure \ref{fig:samplingruntime} for batch sizes between 1 and 100.

While sampling from the numerical optimizer is relatively close to reverse diffusion, 2.5 seconds for TrajDiff and 4.2 seconds for the numerical optimizer, a significant increase in computational efficiency is attained by sampling larger batch sizes from the diffusion model; the diffusion model takes 3.9 seconds for 10 samples, 4.8 seconds for 50 samples, and 45 seconds for 100 samples, compared to 29, 148, and 315 seconds for the numerical trajectory optimizer. A potential future extension of this efficiency is the use a batch of trajectories from TrajDiffuser to warm-start a multi-core implementation of SCvx. Additionally, consider that in terms of non-convex powered descent guidance problems, this formulation is relatively simple; no glideslope, velocity limits, quaternion bounds, angular velocity limits, or angle of attack limits are included in the set of constraints for the model, and only 20 discretization nodes are used. If a problem has additional, more challenging constraints, the runtime of the trajectory optimizer may significantly increase while not affecting the diffusion model's runtime. While other warm-start frameworks would require training a new model to include results from the optimizer, solving the problem with the additional constraints, TrajDiffuser can utilize composition to combine the diffusion model with other models during inference time. An example of this for state-triggered constraints is detailed in the following section.

\subsection{State-Triggered Constraint Enforcement using Compositional Diffusion}

In this experiment, we introduce C-TrajDiffuser: composable trajectory diffusion framework to compose the trained trajectory diffusion model in Section \ref{sec:TrajDiffuser} with an energy function that defines a low energy or log-likelihood when the state-triggered constraint is violated. Our formulation ensures the resulting energy function is differentiable, enabling it to be used as a score function in the composed diffusion model. 

Product composition (Equation \ref{eq: prod}) is used to define the following quantity:

\begin{equation}
    M_{\text{C-trajDiffuser}} = M_{\text{trajDiffuser}} \cap M_{\text{STC}},
\end{equation}

where $M_{\text{C-trajDiffuser}}$ is the product-composed trajectory diffusion model, $M_{\text{trajDiffuser}}$ is the minimum time 6 DoF trajectory diffusion model, and $M_{\text{STC}}$ is the state-triggered constraint energy-based model. The energy function $E_{\bm{\theta}}({\bm{x}}_t)$ for $M_{\text{STC}}$ is formulated as follows:
% 
% \begin{equation}
% E_{\bm{\theta}}({\bm{x}}_t) =
% \begin{cases}
%   0, & \text{if }
%        \begin{aligned}[t]
%        \text{constraint satisfied}\\
%        \end{aligned}
% \\
%   \lambda * \text{violation amount}^2, & \text{otherwise}
% \end{cases}
% \label{eq: statetrigenergy}
% \end{equation}
\begin{equation}
 E_{\bm{\theta}}({\bm{x}}_t) =
\begin{cases}
  0, & \scriptstyle\text{if constraint satisfied} \\
  \lambda \cdot (\text{violation amount})^2, & \scriptstyle\text{otherwise}
\end{cases}
\label{eq:statetrigenergy}
\end{equation}

The formulation in Equation \ref{eq:statetrigenergy} can be used for modeling any inequality constraint, including state-triggered constraints. The following section will show an example of applying this composition framework to a velocity-triggered constraint.

\subsubsection{Product Composition with the Velocity-Triggered Angle of Attack State-Triggered Constraint}

\begin{figure}[htbp]
    \centering
    \includegraphics[width=0.5\textwidth]{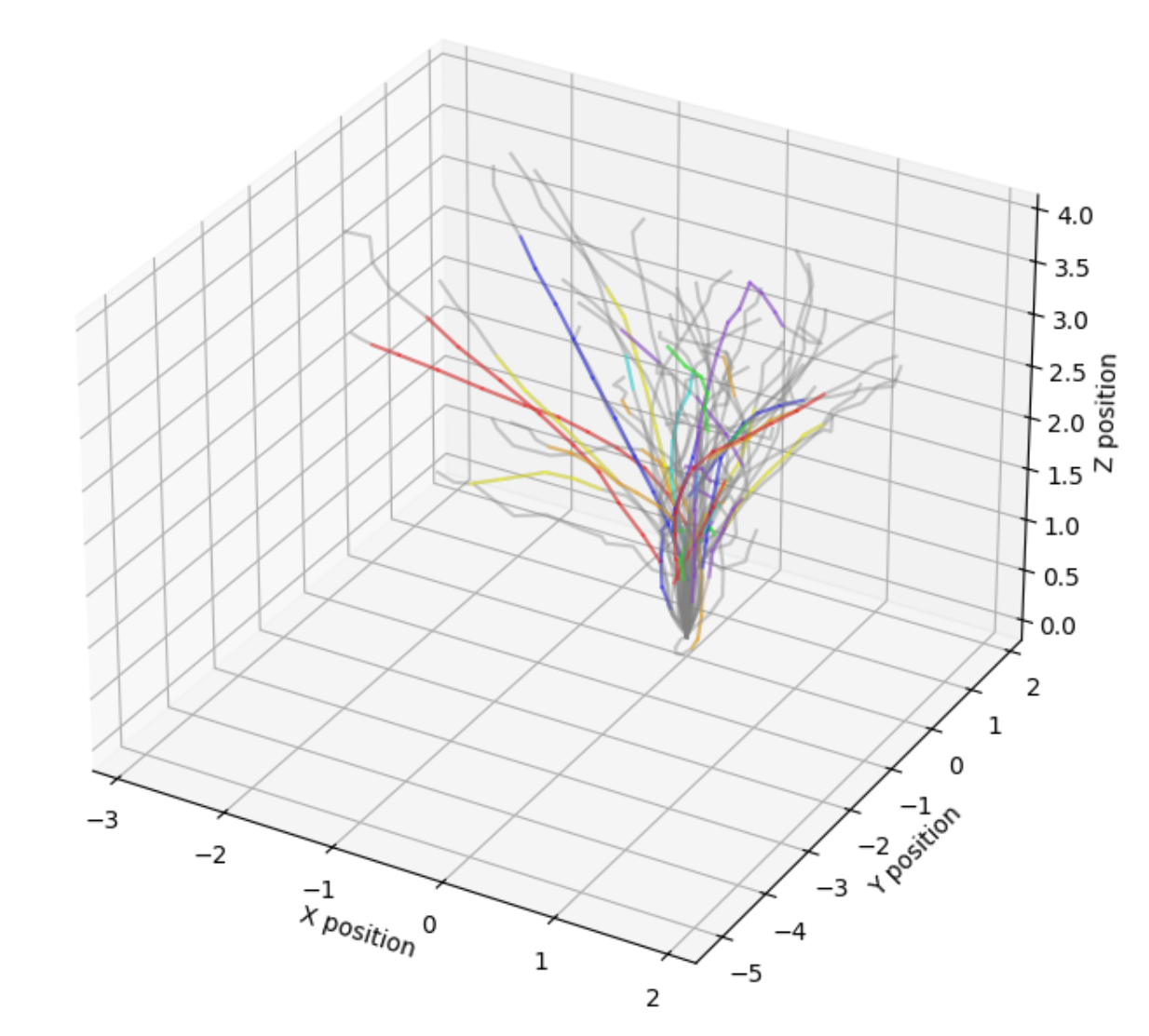}
    \caption{State-triggered constraint satisfaction for 50 trajectory samples without composition. 158 constraints are violated (colored trajectories), resulting in 15.8\% violation.}
    \label{fig:generated_trajectories_position_statetriggered}
\end{figure}

The energy function for the velocity-triggered angle of attack STC is defined by Equation \ref{eq:statetrigenergy} with $\lambda = 10$, a maximum velocity of $V_{\alpha} = 2$, and a maximum angle of attack of $\alpha_{\max} = 10$ degrees. After performing reverse diffusion on the product composition $M_{\text{C-trajDiffuser}}$, the percent of constraints violated can be evaluated. State-triggered constraint satisfaction is evaluated at each time index for 50 trajectories sampled from TrajDiffuser and plotted in Figure \ref{fig:generated_trajectories_position_statetriggered}. Out of the 1000 indices, 158 of them, or 15.8\%, violate the velocity-triggered angle of attack constraint.

\begin{figure}[htbp]
    \centering
    \includegraphics[width=0.5\textwidth]{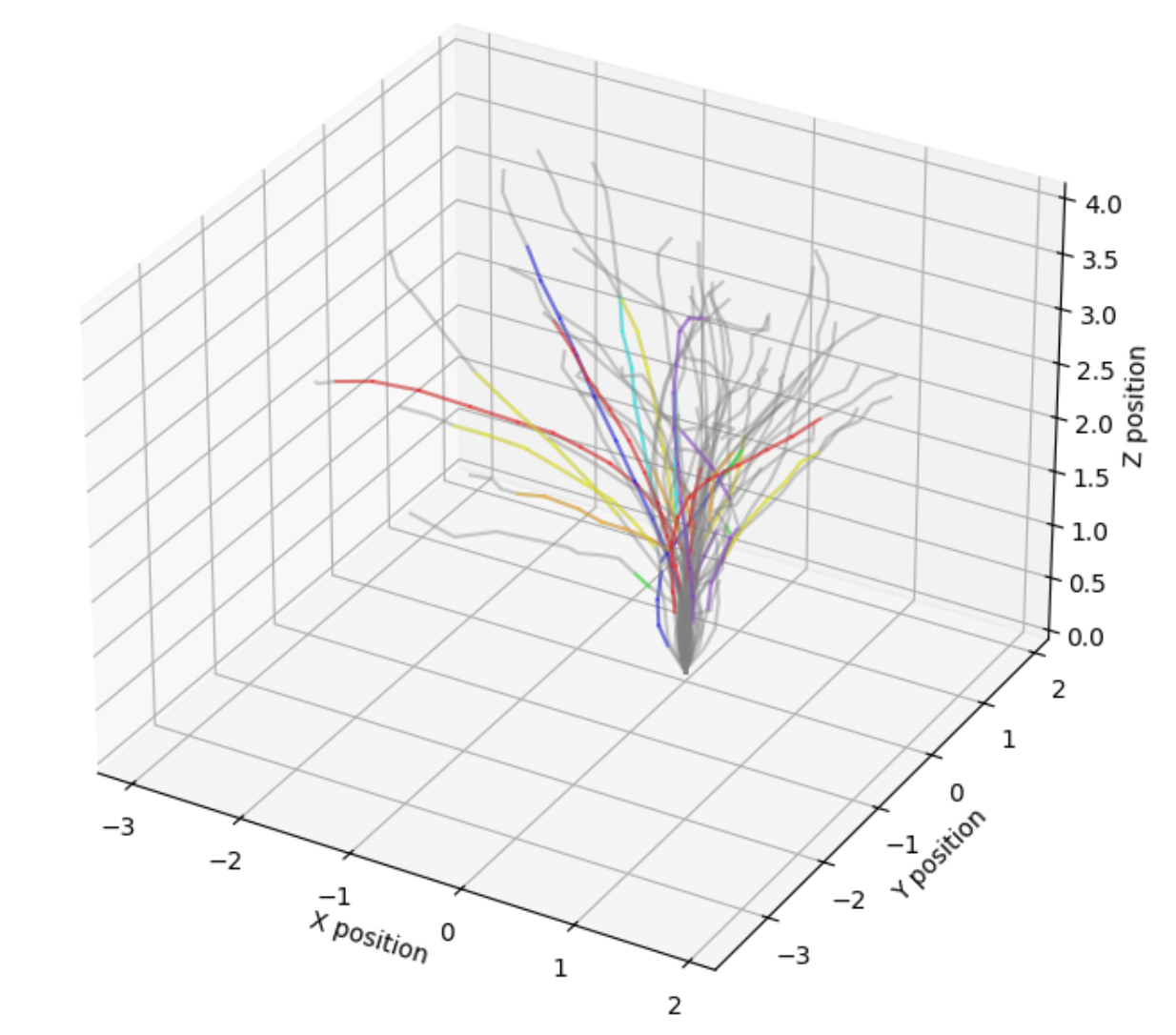}
    \caption{State-triggered constraint satisfaction for 50 trajectory samples with product composition. 127 constraints are violated (colored trajectories), resulting in 12.7\% violation.}
    \label{fig:generated_trajectories_prodcomp_statetriggered}
    
\end{figure}

When product composition C-TrajDiffusion is used to compose the diffusion trajectory model with the energy function defined in Equation \eqref{eq:statetrigenergy}, trajectories following the distribution in Figure \ref{fig:generated_trajectories_prodcomp_statetriggered} are obtained. The same seed for the random number generator and 50 samples were used to ensure any difference in constraint satisfaction is due to the compositional diffusion process. Out of the 1000 time steps, 127 time steps, or 12.7\%, of the timesteps violated the state-triggered constraint.  

Overall, C-TrajDiffuser reduced constraint violation by almost 20 \% compared to TrajDiffuser, which is not trained on the state-triggered constraint. Future work will utilize Markov Chain Monte Carlo (MCMC) sampling to improve constraint satisfaction of compositional models further. 

\subsubsection{Negation Composition with the Velocity-Triggered Angle of Attack State-Triggered Constraint}

This section aims to formulate the negation composition for the velocity-triggered constraint instead of the product composition:

\begin{equation}
    M_{\text{C-trajDiffuser}} = \frac{ M_{\text{trajDiffuser}}} {M_{\text{STC}}},
\end{equation}

where we define $\alpha_0 = 1.3$ is used to scale trajDiffuser and $\alpha_1 = 0.3$ is used to scale
STC. The energy-based could instead be formulated as shown in Equation \ref{eq:statetrigenergyneg}.

\begin{equation}
 E_{\bm{\theta}}({\bm{x}}_t) =
\begin{cases}
  0, & \scriptstyle\text{if constraint is not satisfied} \\
  \lambda \cdot (\text{satisfied amount})^2, & \scriptstyle\text{otherwise}
\end{cases}
\label{eq:statetrigenergyneg}
\end{equation}

where the constraint is satisfaction is defined by the logic,

\begin{equation}
  \neg (g({\bm{z}}) \leq 0) \cup 
 (c({\bm{z}}) \leq 0).
\end{equation}

\begin{figure}[htbp]
    \centering
    \includegraphics[width=0.5\textwidth]{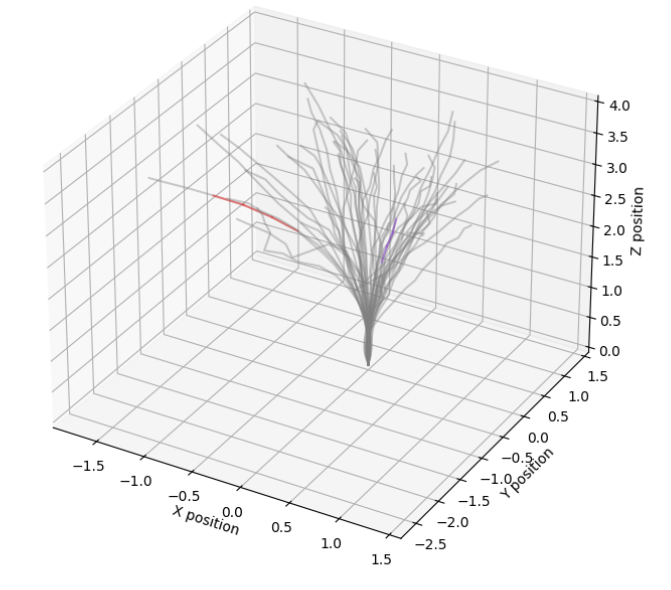}
    \caption{State-triggered constraint satisfaction for 50 trajectory samples with product composition. 6 constraints are violated (colored trajectories), resulting in 0.6\% violation.}
    \label{fig:generated_trajectories_prodcomp_statetriggeredneg}
    
\end{figure}

Figure \ref{fig:generated_trajectories_prodcomp_statetriggeredneg} shows the resulting trajectories after sampling from the negation-composed distribution using reverse sampling. Compared to product C-TrajDiffuser, negation C-TrajDiffuser significantly improves constraint satisfaction; only 0.6\% of time indices violate the velocity-triggered angle of attack constraint.

While product diffusion is most natural when using diffusion models directly, these results demonstrate the benefits of defining an energy-based parameterization; implicitly modeling the score function using the energy function enabled reverse sampling from the negation model.

\subsection{Including Drag using Compositional Diffusion}

\begin{figure*}[htbp]
    \centering
    \includegraphics[width=0.8\textwidth]{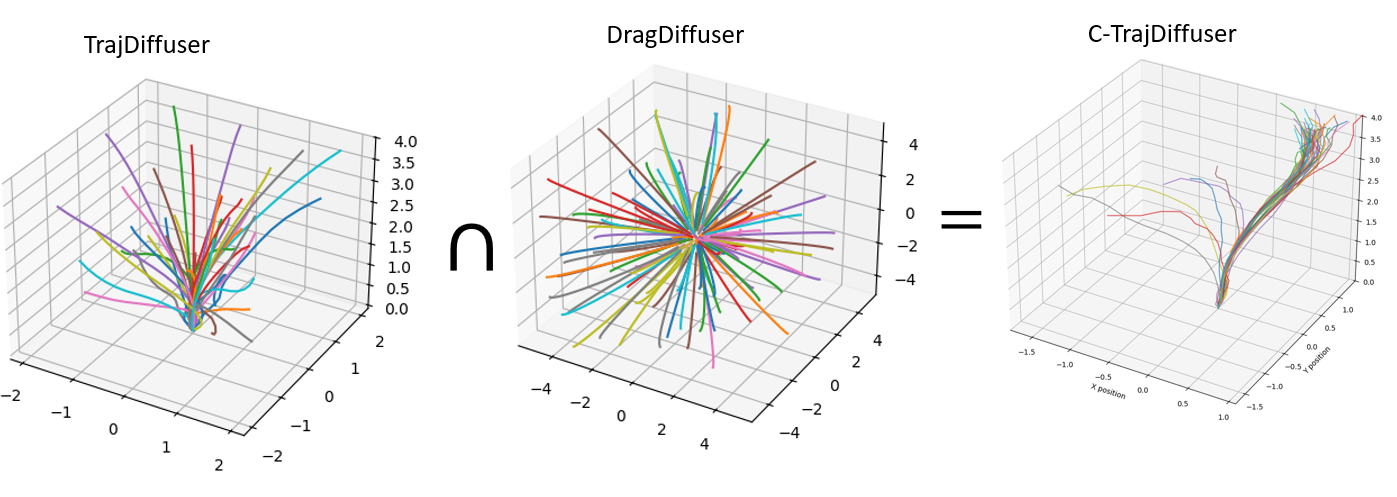}
    \caption{Drag-included trajectories generated by C-TrajDiffuser, composing the 6 DoF trajectory diffusion model, TrajDiffuser, with the 3 DoF diffusion model with drag.}
    \label{fig:drag}
    
\end{figure*}

\begin{figure*}[htbp]
    \centering
    \includegraphics[width=0.8\textwidth]{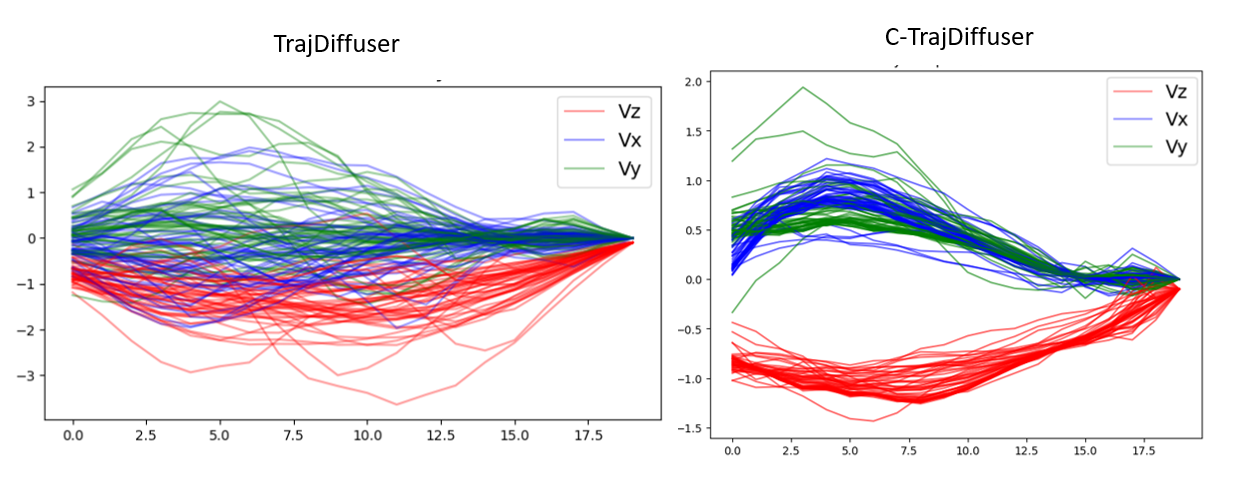}
    \caption{Drag-included velocities generated by C-TrajDiffuser, composing the 6 DoF trajectory diffusion model, TrajDiffuser, with the 3 DoF diffusion model with drag.}
    \label{fig:dragvel}
    
\end{figure*}

While TrajDiffuser does not include drag in the dynamics model, provided a separate diffusion model with dynamics including drag, product composition may be used to obtain trajectories incorporating drag in the trajectory dynamics. Using C-TrajDiffuser from the previous section, we will perform the following product composition:

\begin{equation}
    M_{\text{C-trajDiffuser}} = M_{\text{trajDiffuser}} \cap M_{\text{Drag}},
\end{equation}

where $M_{\text{Drag}}$ is sampled from an easy-to-sample 3 DoF dynamics model:

\begin{align}
    \bm{v}_k &= \bm{v}_{k-1} + \bm{g} + \bm{T}_k - C_d \|\bm{v}_{k-1}\| \bm{v}_{k-1} \Delta t \\
    \bm{r}_k &= \bm{r}_{k-1} + \bm{v}_k \Delta t,
\end{align}

and a simple proportional derivative control input:

\begin{equation}
    \bm{T}_k = K_p (\bm{r}_\text{target} - \bm{r}_{k-1}) + K_d (\bm{v}_\text{target} - \bm{v}_{k-1}),
\end{equation}

where all parameters are the same as those defined in Table \ref{tab:algorithm_settings} except for the drag coefficient $C_d = 0.1$ and the PD control gains $K_p = 5$, $K_d = 2$. While this is a 3 DoF model, which does not include the quaternions or angular velocities, these parameters can be zeroed out such that when product diffusion is taken (Equation \ref{eq: prod}), they are only determined by the product contribution by TrajDiffuser. Figure \ref{fig:drag} shows trajectory samples from the composed distribution, and Figure \ref{fig:dragvel} compares the TrajDiffuser velocities to the C-TrajDiffuser velocities.

\begin{figure}[htbp]
    \centering
    \includegraphics[width=0.5\textwidth]{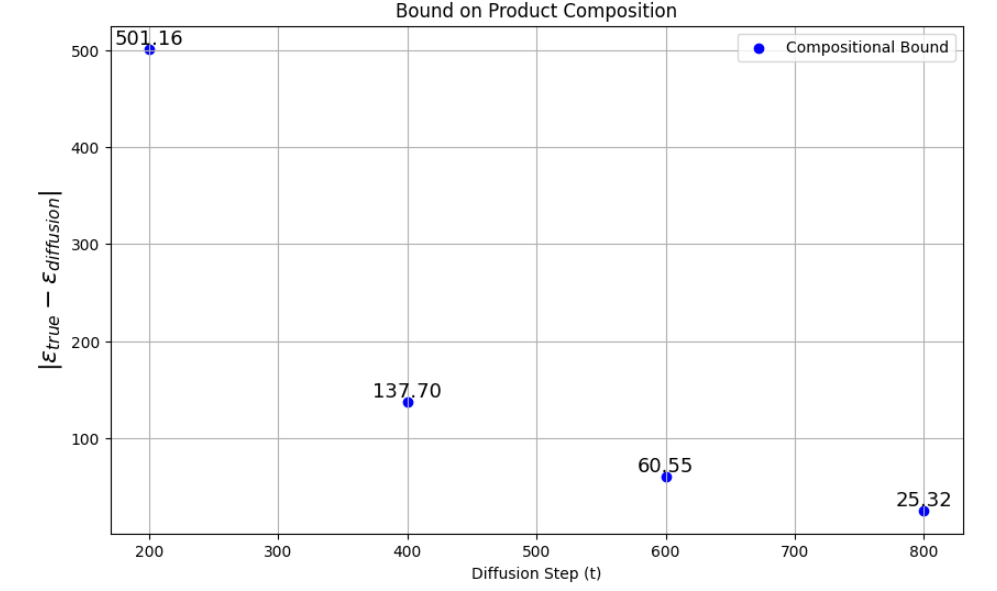}
    \caption{The bound on score function error plotted over 200, 400, 600, and 800 diffusion timesteps.}
    \label{fig:boundplot}
    
\end{figure}

Generally, drag reduces the spacecraft's travel distance and decreases its maximum velocity in the unactuated case since drag acts in the direction of opposing motion. Observing Figures \ref{fig:drag} and \ref{fig:dragvel}, compared to Figures \ref{fig:generated_trajectories_position} and \ref{fig:dragvel} from TrajDiffuser, we see a decrease in both xy distance and velocity magnitudes, indicating the composition process successfully biased the generated trajectories towards incorporating drag in the dynamics model. The derived Equation \ref{eq: bound} can be used to determine an upper limit on the product diffusion model error, guaranteeing the accuracy of using reverse diffusion for compositional sampling. The upper bound for this product composition is plotted in Figure \ref{fig:boundplot}, where $N = 2$ composed models, $T=1000$ diffusion steps, $d=2$ for the dimensionality of the dataset, the trace of the $x_0$ variance is computed by taking the trace of the variance from a sample of 50 trajectories from the numerical optimizer, and the trace of the $x_0$ variance over $q_0$ is computed using the trace of the variance for 50 samples from TrajDiffuser.

While the bound is relatively tight as the diffusion process approaches ${x_T} {\sim} {\mathcal} (0, I)$, it gets looser as $T \rightarrow 0$. Future work includes obtaining a tighter bound for low diffusion time steps and computing additional bounds for mixture and negation with reverse diffusion.

%\section{Discussion}
%\label{sec:discussion}

\section{Conclusions}
\label{sec:conclusions}

This work formulates the first diffusion model for powered descent guidance, TrajDiffuser. The model includes the state and control for 6-degree-of-freedom dynamics. Experiments evaluate the model's constraint satisfaction, optimality, time complexity, and composition performance. TrajDiffuser-sampled trajectories were close to locally optimal solutions generated using numerical optimization. Computational benchmarking reveals that TrajDiffuser reduces runtimes by up to 86\% for 100 samples when compared to numerical optimization. A novel bound on score function error is derived to estimate the worst-case performance for product composition. Lastly, compositional trajectory diffusion, C-TrajDiffuser, is introduced. C-TrajDiffuser was tested on both product and negation composition for state-triggered constraints and product diffusion for adding drag into the trajectory dynamics model. C-TrajDiffuser resulted in a 96\% reduction in constraint violation for negation composition compared to TrajDiffuser alone. This work represents the first step towards a compositional trajectory generation toolbox for generalizable and efficient trajectory generation.

\appendices{}              % note there is no {} to put a title. Each appendix has its own title
%%%%%%%%%%%%%%%%%%%%%%%%%%%%%%%%%%%%%%%%%%%%%%%%%%%%%%%%%%%%%%%%%%%%%%%%%%%%%%%%%%%%%%%%%%%%%%%%%
% For a single appendix, use the \appendix{} keyword and do not use the \section command.

\section{Reverse Diffusion Sampling Bounds}\label{sec: appendix product}        % first appendix
%%%%%%%%%%%%%%%%%%%%%%%%%%

\subsection{Product Composition}

The following derivation computes the difference between the reverse diffusion-defined product composition score function and the true product composition score function. From the definition of the reverse diffusion-defined product composition (Equation \eqref{eq: prod}), the score function is equivalent to
\begin{align}
    {\bm{\epsilon}}_{{\bm{\theta}} \; \text{prod}} (x,t) &= \sum_{i=1}^N {\bm{\epsilon}}_{\bm{\theta}}^i \\
    &= \sum_{i=1}^N \nabla \log q^i (x_t), \\
\end{align}
since $q^i(x_t) = \int p_{\bm{\theta}}^i (x_0) q(x_t | x_0) dx_0$. Each score function for the composed models can be computed as
\begin{align}
    \nabla \log q^i (x_t) &= - \frac{1}{1-\bar \alpha_t} (x_t - \sqrt{\bar \alpha_t} {\mathbb{E}}_{q^i (x_0 | x_t)} [x_0]).
\end{align}
Therefore adding the score functions for every composed model results in the following expression:
\begin{align}
    {\bm{\epsilon}}_{{\bm{\theta}} \; \text{prod}} (x,t) &= \sum_{i=1}^N - \frac{1}{1-\bar \alpha_t} (x_t - \sqrt{\bar \alpha_t} {\mathbb{E}}_{q^i (x_0 | x_t)} [x_0]) \\
    &= - \frac{1}{1-\bar \alpha_t} (N x_t - \sqrt{\bar \alpha_t} \sum_{i=1}^N {\mathbb{E}}_{q^i (x_0 | x_t)} [x_0]).
\end{align}
Similarly, the product distribution's score function is,
\begin{align}
    \tilde {\bm{\epsilon}}_{{\bm{\theta}} \; \text{prod}} (x,t) &= - \frac{1}{1-\bar \alpha_t} (x_t - \sqrt{\bar \alpha_t} {\mathbb{E}}_{\tilde q_{\text{prod}} (x_0 | x_t)} [x_0]).
\end{align}
Computing the difference in score functions gives us
\begin{align}
    \delta_{\text{prod}} (x,t) &= |{\bm{\epsilon}}_{{\bm{\theta}} \; \text{prod}} (x,t) - \tilde {\bm{\epsilon}}_{{\bm{\theta}} \; \text{prod}} (x,t)| \\
    &= \frac{1}{1-\bar \alpha_t} |(N-1) x_t - \sqrt{\bar \alpha_t} (\sum_{i=1}^N {\mathbb{E}}_{q^i (x_0 | x_t)} [x_0] \\
    &- {\mathbb{E}}_{\tilde q_{\text{prod}} (x_0 | x_t)} [x_0])|.
\end{align}
If we define the average of the composed models as $\bar {\bm{\mu}} = \frac{1}{N} \sum_{i=1}^N {\mathbb{E}}_{q^i (x_0 | x_t)} [x_0]$, then we can factor out the number of composed models $N$ to get the difference in means:
\begin{equation}
    \Delta {\bm{\mu}} = N ( {\mathbb{E}}_{\tilde q_{\text{prod}} (x_0 | x_t)} [x_0] - \bar {\bm{\mu}}).
\end{equation}
Plugging this in, we have,
\begin{align}
    \delta_{\text{prod}} (x,t) 
    &= \frac{1}{1-\bar \alpha_t} |(N-1) x_t + \sqrt{\bar \alpha_t} \Delta {\bm{\mu}}|.
\end{align}
To derive a bound for $\delta_{\text{prod}} (x,t)$, we need to bound the diffusion sample $x_t$ and the difference in means $\Delta {\bm{\mu}}$. Using the variance of the diffusion process $\text{Var}(x_t) = \bar \alpha_t \text{Var} (x_0) + (1-\bar{\alpha}_t) I$, $\|x_t\|$ is bounded as long as $\text{Var} (x_0)$ is bounded. By applying the triangle inequality $|(N-1) x_t + \sqrt{\bar \alpha} \Delta {\bm{\mu}}| \leq (N-1)\|x_t\| + \sqrt{\bar \alpha_t} \| \Delta {\bm{\mu}} \|$, we can bound $\delta_{\text{prod}} (x,t)$ as,
\begin{equation}
    \delta_{\text{prod}} (x,t) 
    \leq \frac{1}{1-\bar \alpha_t} \left((N-1)\|x_t\| + \sqrt{\bar \alpha_t} \| \Delta {\bm{\mu}} \|\right).
\end{equation}
Finally, we can sum over $T$ diffusion steps to get the final bound for the full reverse process:
\begin{equation}
    \delta_{\text{prod}} (x) 
    \leq \sum_{t=1}^T \frac{1}{1-\bar \alpha_t} \left((N-1)\|x_t\| + \sqrt{\bar \alpha_t} \| \Delta {\bm{\mu}} \|\right).
\end{equation}
To ensure $\|x_t\|$ and $\| \Delta {\bm{\mu}} \|$ are bounded, we can consider the law of total variance;
${{{\text{Var}}_{{q_{\text{prod}}}}}} (x_0) \leq {{\text{Var}}_{q^i}} (x_0)$ for any composed model $i \in 1, \dots, N$. Applying the Cauchy-Schwarz inequality,
\begin{align}
    \| {\bm{\mu}}_{\text{prod}} - {\bm{\mu}}_i \| \leq \sqrt{{{\text{Var}}_{q_{\text{prod}}}} (x_0) + {{\text{Var}}_{q^i}} (x_0)}.
\end{align}
Since we have ${{\text{Var}}_{{q_{\text{prod}}}}} (x_0) \leq {{\text{Var}}_{q^i}} (x_0)$, we know that ${{\text{Var}}_{q_{\text{prod}}}} (x_0) + {{\text{Var}}_{q^i}} (x_0) \leq 2 {{\text{Var}}_{q^i}} (x_0)$, giving us the following bound:
\begin{align}
    \| {{\bm{\mu}}_{\text{prod}}} - {\bm{\mu}}_i \| \leq \sqrt{ 2 {{\text{Var}}_{q^i}} (x_0)}.
\end{align}
Now, computing the expression for $\| \Delta {\bm{\mu}} \|$, we have
\begin{align}
    \| \Delta {\bm{\mu}} \| &= N \| {\bm{\mu}}_{\text{prod}} - \bar {\bm{\mu}} \| \\
    &\leq \sum_{i=1}^N \sqrt{ 2 {\text{Var}}_{q^i} (x_0)}.
\end{align}
If we either assume all variances are equal across models, ${{\text{Var}}_{q^i}} (x_0) = {{\text{Var}}_{q^j}} (x_0)$ for $i, j \in 1, \dots N$, or bound using only the maximum variance of all the composed models, we can simplify the bound to be
\begin{align}
    \| \Delta {\bm{\mu}} \| &\leq N \sqrt{ 2 {\text{Var}}_{q^i} (x_0)}, \\
    &\leq N \sqrt{ 2 {\text{Tr}} ( {\text{Var}}_{q^i} (x_0))},
\end{align}
where the trace operator sums the variances along each dimension of $x_0$. To bound the covariance matrix of $x_t$, we can use the forward process computation, $\text{Var}(x_t) = \bar \alpha_t \text{Var} (x_0) + (1-\bar{\alpha}_t) I$. For a Gaussian random variable $x_t$,
\begin{align}
    {\mathbb{E}} \|x_t\|^2 &= {\text{Tr}} ( {\textbf{Var}} (x_t)) + \| {\mathbb{E}} [x_t] \|^2.
\end{align}
If $x_0$ is normalized with a mean of zero ${\mathbb{E}} [x_0] = 0$, then ${\mathbb{E}} [x_t] = 0$ as well. Plugging in $\text{Var}(x_t)$, we have
\begin{align}
    {\mathbb{E}} \|x_t\|^2 &= {\text{Tr}} ( \bar \alpha_t {\text{Var}} (x_0) + (1-\bar{\alpha}_t) I).
\end{align}
Finally, bounding $\| x_t \|$ with the expected norm and taking the sum of the variances along each dimension using the trace operator,
\begin{align}
    \|x_t\| \leq {\sqrt{{\mathbb{E}} \|x_t\|^2}} &= \sqrt{{\text{Tr}} ( \bar \alpha_t \text{Var} (x_0) + (1-\bar{\alpha}_t) I)} \\
    &= \sqrt{\bar \alpha_t {\text{Tr}} (\text{Var} (x_0)) + (1-\bar{\alpha})_t) d},
\end{align}
where $d$ is the dimensionality of $x_t$.

%%%%%%%%%%%%%%%%%%%%%%%%%%%%%%%%%%%%%%%%%%%%%%%%%%%%%%%%%%%%%%%%%%%%%%%%%%%%%%%%%%%%%%%%%%%%%%%%%%%%%%
\acknowledgments
This work was supported in part by a NASA Space Technology Graduate Research Opportunity 80NSSC21K1301.

%%%%%%%%%%%%%%%%%%%%%%%%%%%%%%%%%%%%%%%%%%%%%%%%%%%%%%%%%%%%%%%%%%%%%%%%%%%%%%%%%%%%%%%%%%%%%%%%%%%%%%
\bibliographystyle{IEEEtran}
%\bibliography{references}
\bibliography{references}

%%%%%%%%%%%%%%%%%%%%%%%%%%%%%%%%%%%%%%%%%%%%%%%%%%%%%%%%%%%%%%%%%%%%%%%%%%%%%%%%%%%%%%%%%%%%%%%%%%%%%%
\thebiography
%% This biostyle allows you to insert your photo size 1in X 1.25in

\begin{biographywithpic}
{Julia Briden}{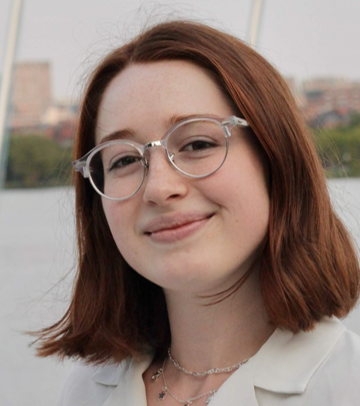}
received her B.S. and Master's degrees from the Illinois Institute of Technology in 2021 and her Ph.D. in Aeronautics and Astronautics from MIT in 2024. As a NASA Space Technology Graduate Research Opportunities (NSTGRO) fellow at the MIT ARCLab, she has worked to develop computationally efficient algorithms for spacecraft entry, descent, and landing and accurate space weather prediction and modeling methods for resident space object reentry. Previously, she worked as a Robotics, Guidance, and Control Intern at NASA JPL in 2023, as a Flight Mechanics and Trajectory Design Intern at NASA JSC in 2022, as a GN\&C Intern for New Glenn hardware-in-the-loop (HIL) at Blue Origin in 2021, and as a Flight Systems Engineering Intern for the NISAR mission at NASA JPL in 2020.

\end{biographywithpic}

\begin{biographywithpic}
{Yilun Du}{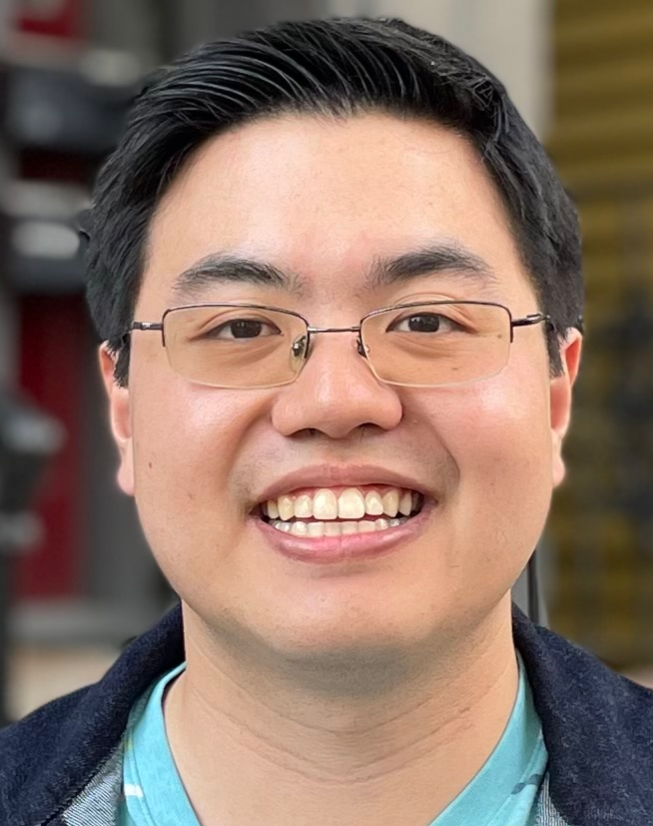} received his B.S. degree from MIT in 2019, Master's degree from MIT in 2020, and his Ph.D in Electrical Engineering and Computer Science in 2024.  He is currently a senior research scientist at Google Deepmind and starting July 2025, he will be an Assistant Professor at Harvard in Kempner Institute and Computer Science. 
\end{biographywithpic} 

\begin{biographywithpic}
{Enrico Zucchelli}{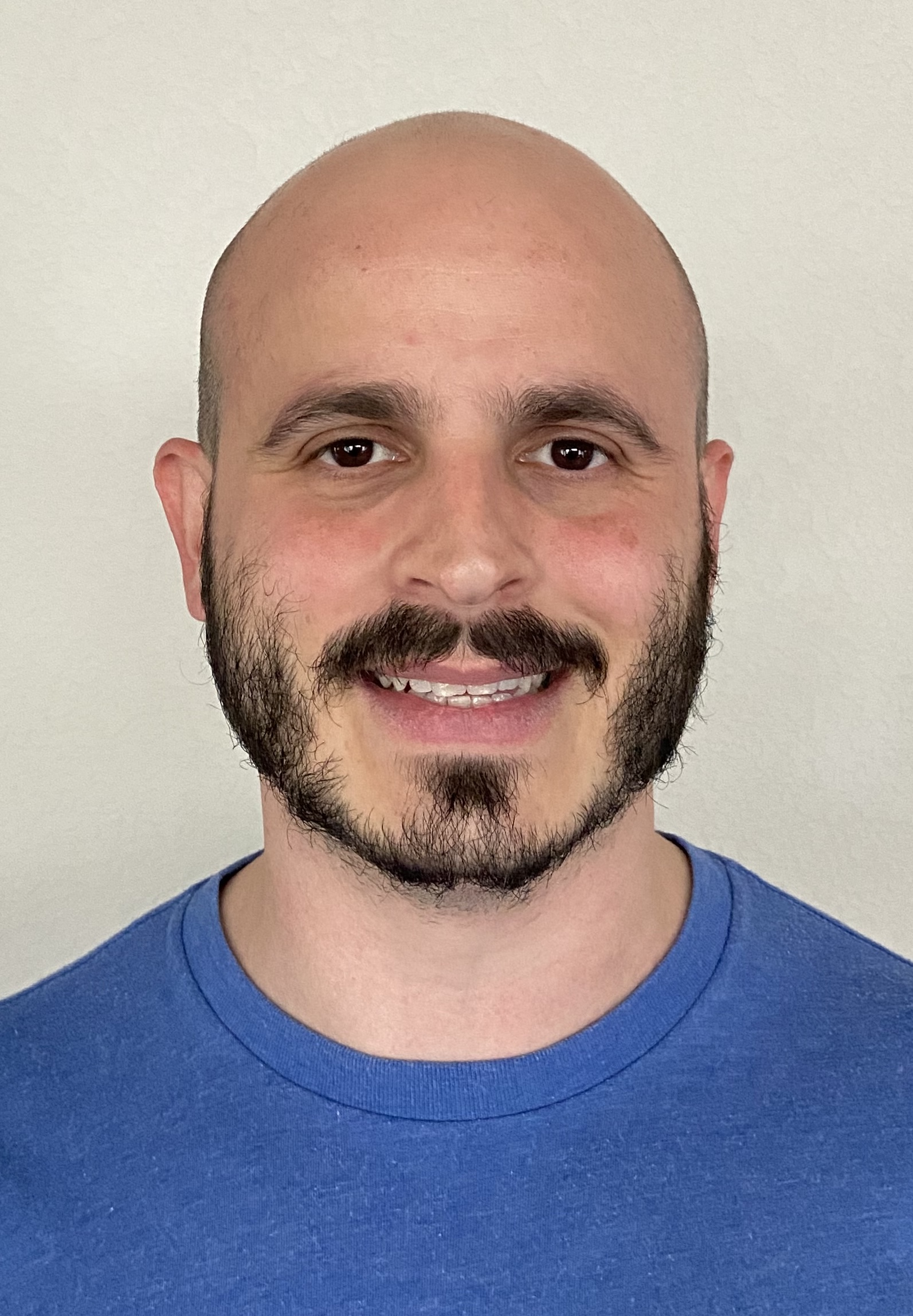} received his B.S. degree from Politecnico di Milano, Italy, his M.S. degree from Delft University of Technology, The Netherlands, and his Ph.D. degree in Aerospace Engineering from The University of Texas at Austin. Since February 2024 he is a Postdoctoral Associate at MIT's Department of Aeronautics and Astronautics.
\end{biographywithpic}

\begin{biographywithpic}
{Richard Linares}{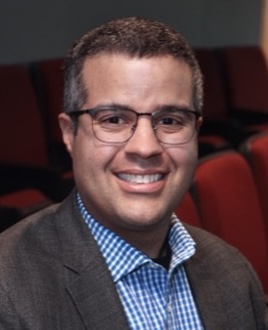} joined MIT’s Department of Aeronautics and Astronautics as an Assistant Professor in July of 2018. Before joining MIT, he was an Assistant Professor at the University of Minnesota’s Aerospace Engineering and Mechanics Department. He received his B.S., M.S. and Ph.D. degrees in aerospace engineering from University at Buffalo, The State University of New York. He was a Director’s Postdoctoral Fellow at Los Alamos National Laboratory and also held a postdoctoral associate appointment at the United States Naval Observatory. His research areas are astrodynamics, estimation and controls, satellite guidance and navigation, space situational awareness, and space traffic management. Richard Linares is a recipient of the AFOSR Young Investigator Research Program Award in 2018 and the 2020 DARPA Young Faculty Award.
\end{biographywithpic}

\end{document}